\documentclass{ieeeaccess}
\usepackage{cite}
\usepackage{amsmath,amssymb,amsfonts}

\usepackage{bm}

\usepackage[hyphens]{url}
\usepackage[table]{xcolor}
\usepackage{hyperref}
\hypersetup{
    colorlinks=false,
    citebordercolor=white,
    linkbordercolor=white,
    urlbordercolor=cyan,
}
\usepackage{graphicx}
\usepackage{lipsum}
\usepackage{pifont}
\usepackage{nccmath}
\usepackage{comment}
\usepackage{balance}
\usepackage{multirow}
\usepackage{textcomp}
\usepackage{multirow}
\usepackage{subcaption}
\usepackage{booktabs}
\usepackage{siunitx}
\usepackage{tabularx}
\usepackage{algorithm}
\usepackage{dblfloatfix}

\usepackage[noend]{algpseudocode}

\makeatletter
\AtBeginDocument{\DeclareMathVersion{bold}
\SetSymbolFont{operators}{bold}{T1}{times}{b}{n}
\SetSymbolFont{NewLetters}{bold}{T1}{times}{b}{it}
\SetMathAlphabet{\mathrm}{bold}{T1}{times}{b}{n}
\SetMathAlphabet{\mathit}{bold}{T1}{times}{b}{it}
\SetMathAlphabet{\mathbf}{bold}{T1}{times}{b}{n}
\SetMathAlphabet{\mathtt}{bold}{OT1}{pcr}{b}{n}
\SetSymbolFont{symbols}{bold}{OMS}{cmsy}{b}{n}
\renewcommand\boldmath{\@nomath\boldmath\mathversion{bold}}}
\makeatother

\def\BibTeX{{\rm B\kern-.05em{\sc i\kern-.025em b}\kern-.08em
    T\kern-.1667em\lower.7ex\hbox{E}\kern-.125emX}}

\begin{document}
\history{Accepted 12 September 2024}
\doi{10.1109/ACCESS.2024.0429000}

\title{Tightly-Coupled LiDAR-IMU-Wheel Odometry with Online Calibration of a Kinematic Model for Skid-Steering Robots}
\author{\uppercase{Taku Okawara}\authorrefmark{1},
\uppercase{Kenji Koide}\authorrefmark{2}, 
\uppercase{Shuji Oishi}\authorrefmark{2}, 
\uppercase{Masashi Yokozuka}\authorrefmark{2}, \\
\uppercase{Atsuhiko Banno}\authorrefmark{2}, 
\uppercase{Kentaro Uno}\authorrefmark{1}, 
and \uppercase{Kazuya Yoshida}\authorrefmark{1}}

\address[1]{The Space Robotics Lab. in the Department of Aerospace Engineering, Graduate School of Engineering, Tohoku University, Sendai, Miyagi, Japan}
\address[2]{The Department of Information Technology and Human Factors, the National Institute of Advanced Industrial Science and Technology, Tsukuba, Ibaraki, Japan}
\tfootnote{This work was supported in part by JSPS KAKENHI Grant Number 22KJ0292 and a project commissioned by the New Energy and Industrial Technology Development Organization (NEDO).}

\markboth
{T. Okawara et.al.: Tightly-Coupled LiDAR-IMU-Wheel Odometry with Online Calibration of a Kinematic Model for Skid-Steering Robots}
{Taku Okawara et.al.: Tightly-Coupled LiDAR-IMU-Wheel Odometry with Online Calibration of a Kinematic Model for Skid-Steering Robots}

\corresp{Corresponding author: Taku Okawara (e-mail: okawara.taku.t3@dc.tohoku.ac.jp).}

\begin{abstract}
Tunnels and long corridors are challenging environments for LiDAR-based odometry estimation algorithms because a LiDAR point cloud should degenerate (i.e., point cloud matching cannot work properly) in such environments. 
To tackle point cloud degeneration, this study presents a tightly-coupled LiDAR-IMU-wheel odometry algorithm incorporating online calibration of a kinematic model for skid-steering robots. 
We propose a \textit{full linear wheel odometry factor}, which not only serves as a motion constraint but also performs the online calibration of kinematic models. 
Despite the dynamically changing kinematic parameters (e.g., wheel radii changes caused by tire pressures) and terrain conditions, our method can address the model error via online calibration. 
Moreover, our method enables an accurate localization in cases of degenerated environments, such as long and straight corridors, by calibration while point cloud-based constraints sufficiently operate. 
Furthermore, we estimate the uncertainty (i.e., covariance matrix) of the wheel odometry online for creating a constraint with a reasonable statistical model even in rough terrains. 
The proposed method is validated through three experiments.
The first indoor experiment shows that the proposed method is robust in severe degeneracy cases (long corridors) and changes in the wheel radii. 
The second outdoor experiment demonstrates that our method accurately estimates the sensor trajectory despite rough outdoor terrain thanks to online uncertainty estimation of wheel odometry.
The third experiment shows the proposed online calibration enables robust odometry estimation in a condition that terrains change.
\end{abstract}

\begin{keywords}
Odometry, Factor graph optimization, Sensor fusion, Online calibration, Point cloud degeneration, Online uncertainty (covariance matrix) estimation, State estimation, Wheel slippage
\end{keywords}

\titlepgskip=-21pt

\maketitle

\section{Introduction}

\begin{figure*}[tb]
  \centering
  \includegraphics[width=1\linewidth]{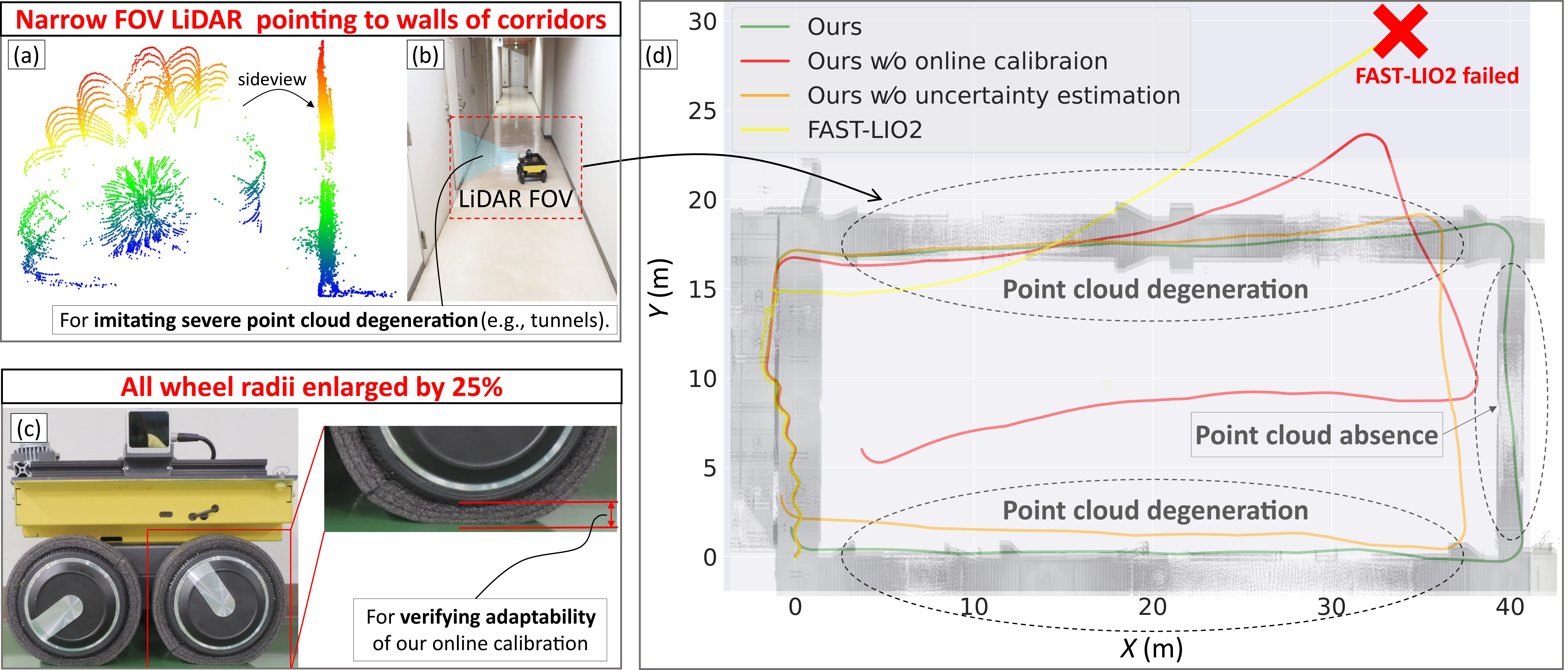}
  \caption{Indoor experiment conditions and odometry estimation results: (a) Degenerated point clouds were obtained in long-term because (b) the narrow FOV LiDAR (Livox AVIA) pointed to walls of long straight corridors for imitating severe point cloud degeneration such as tunnels.
  (c) All wheel radii of the skid-steering were enlarged by $25~\%$ (i.e., large model error) for verifying adaptability of our online calibration method.
  (d) The proposed method (Ours) was the most accurate of all cases thanks to online calibration and online uncertainty estimation although this long straight corridor includes the two areas where point clouds were degenerated and unavailable.
  The robot traveled about \SI{120}{m}, \SI{302}{s}.
  Therefore, the proposed method can tackle long-term point cloud degeneration, point cloud absence, and large kinematic model errors.}
  \label{fig:exp1_traj_and_experimental_conditions}
\end{figure*}

An accurate and robust odometry estimation is critical for autonomous robots to achieve reliable navigation and mapping in real-time. The state-of-the-art LiDAR odometry~\cite{liu2021balm} and LiDAR IMU odometry~\cite{qin2020lins, xu2022fast} can accurately estimate the robot pose thanks to the tight fusion of LiDAR and IMU data in feature-rich environments where point cloud matching works well. 
However, the point cloud-based constraints can corrupt in featureless environments where the LiDAR point clouds degenerate in long-term (e.g., long corridors and tunnels); thus such environments are a challenge of LiDAR-IMU odometry. 
Although IMU measurements can help in short-term, IMU-based estimations accumulate drift and are corrupted by the long-term degeneration of LiDAR data. 
Compared to IMU that needs double integration of an accelerometer to a sensor translational displacement, wheel encoders measure the wheel angular velocities requiring only a single integration, which results in a slower accumulation of errors; thus sensor fusion of the LiDAR, IMU, and wheel encoders can accomplish an accurate odometry estimation that is robust against the long-term degeneration of point clouds.
Among wheel robots, skid-steering robots are widely used in various fields thanks to their high road ability and mechanical simplicity. Skid-steering robots rotate by skidding their wheels based on different angular velocities of the left and right wheels, instead of using a steering mechanism. This mechanical simplicity makes the wheel odometry estimation more challenging compared to differential-drive robots because skid-steering robots' wheel odometry must consider terrain-dependent parameters (e.g.,~longitudinal and lateral slip ratio), the robot's center of mass, among others~\cite{anousaki2004dead}. 
To conduct accurate wheel odometry estimation, these parameters must be estimated online as the parameters dynamically change based on the terrain and wheel conditions.
Hence, some works calibrated the parameters to obtain more accurate wheel odometry models and enhance robustness of the localization by fusing other exteroceptive sensors (e.g., camera~\cite{lee2020visual, zuo2019visual, zuo2022visual}, LiDAR~\cite{kummerle2011simultaneous, kummerle2011simultaneous_journal, bedin2023teach})--based constraints.
The LiDAR-based and camera-based constraints are not significantly different for the theory of online calibration.
While a camera must estimate a depth to create constraints, LiDAR can measure a depth directly and accurately; thus LiDAR-based calibration can be more accurate and stable than a camera-based one.
Although LiDAR has traditionally been considered highly expensive compared to cameras, recent advancements have led to the availability of affordable and high-performance LiDARs.
Therefore, we apply LiDAR-based constraints to the proposed method.

In this study, we propose a tightly coupled LiDAR-IMU-wheel odometry method with the online calibration of the kinematic parameters of wheel robots. This is formulated as a factor graph optimization problem. 
We especially tackle the online calibration for skid-steering robots as a more challenging problem. As a general wheel odometry model, we used a full linear model that the velocity of the robot is assumed to be proportional to the angular velocity of the wheels. The online calibration of the full linear model can implicitly deal with unknown kinematic parameters and environments which is difficult to model rigorously. Therefore, we propose a \textit{full linear wheel odometry factor} and jointly optimize the sensor trajectory and kinematic parameters online. Furthermore, we estimate the uncertainty (i.e., covariance matrix) of wheel odometry according to the roughness of the terrain and incorporate the uncertainty into the full linear wheel odometry factor. For example, for a flat terrain, the uncertainties along the height, roll, and pitch direction are estimated to be low, whereas for an uneven terrain, these uncertainties are large.

The main contributions of this study are three-fold:
\begin{enumerate}
  \item We proposed a tightly coupled LiDAR-IMU-wheel odometry to deal with severe degeneration of point clouds and the kinematic model errors of skid-steering robots. We jointly optimized the sensor trajectory and kinematic parameters, which depend on directly-non-observable phenomena or values (e.g., wheel slippage and kinematic model errors), using the full linear wheel odometry factor.
  \item We explicitly estimated the covariance matrix for the full linear wheel odometry factor as a 6~DoF motion. Experimental results validate that it can adapt the reliability of the full linear wheel odometry factor, resulting in accurate localization across environments with different ground surface conditions.
  \item We publish a code of the full linear wheel odometry factor as an open source. \href{https://github.com/TakuOkawara/full_linear_wheel_odometry_factor}{https://onl.la/Asn33BE}
\end{enumerate}

\section{Related Work}

\subsection{Offline calibration for parameters of skid-steering robots}
The wheel odometry computation of skid-steering robots involves environment-dependent parameters and is complicated; thus, this problem has been addressed through offline calibration in many studies. For example, considering the extended differential drive~\cite{mandow2007experimental}, instantaneous centers of rotation (ICR) parameters were introduced to better model the motion of skid-steering robots by extending the ideal differential drive model. The radius of curvature (ROC)-based kinematic model~\cite{wang2015analysis} was experimentally derived based on the relationship between the ICR parameter and ROC of the robot motion. The full linear model~\cite{anousaki2004dead} is a generic expression that describes the relationship between the velocity of the robot and the angular velocity of the wheels.

Furthermore, these methods have been evaluated in different environments, such as a flat plane and uneven terrain~\cite{baril2020evaluation}. The results demonstrated that the full linear model is more accurate than the other methods thanks to its flexibility according to the vehicle configurations and terrain conditions. Notably, whereas these studies were conducted offline, the proposed method addresses online calibration.

\subsection{Online calibration for wheel robots}\label{subsection: senction22}
In the case of wheel robots with a differential drive, some studies have conducted online calibration of kinematic parameters. 
Simultaneous Calibration, Localization, and Mapping~\cite{kummerle2011simultaneous, kummerle2011simultaneous_journal} were proposed to calibrate internal parameters (i.e., wheel radius and wheelbase) and simultaneously localize the robot pose. 
These methods applied a graph-based SLAM framework to the online calibration problem. 
In addition to the internal parameters, the extrinsic parameters between a robot frame and an IMU frame, and time offsets between wheel encoders and the IMU were incorporated into a Kalman filter-based visual-inertial odometry with online calibration~\cite{lee2020visual}.
However, these approaches are based on the differential drive robots, assuming the ideal condition that wheel slippage does not occur.

Some works calculate the uncertainty of wheel odometry estimation by propagating an uncertainty of kinematic parameters and wheel encoder measurements~\cite{lee2020visual, bedin2023teach}.
While these approaches calculate only the uncertainty of 2D motion for wheel odometry in principle, we estimate the uncertainty in 3D by incorporating LiDAR and IMU measurements.

Zhao et.al. conducted visual-IMU-wheel odometry with online calibration of extrinsic parameters and extended this work to LiDAR-IMU-wheel odometry (\href{https://github.com/zhh2005757/FAST-LIO-Multi-Sensor-Fusion}{FAST-LIO-Multi-Sensor-Fusion}~\cite{zhao2023vehicle}). This work incorporates only longitudinal velocity by wheel encoders in contrast to the proposed method, which also considers lateral and rotational motion constraints and uncertainty of six~degrees-of-freedom~(DoF) motion. Furthermore, the longitudinal velocity was directly used to define the constraint instead of a kinematic model; thus wheel encoder-based constraint is fused by loosely coupling way. This algorithm was verified by a skid-steering robot; however, severely featureless environments were not included in the evaluation condition.
The most relevant studies are~\cite{zuo2019visual, zuo2022visual}, which conducted vision-based online calibration of kinematic models for skid-steering robots. 
These methods demonstrated that a fusing of vision and IMU, and ICR-based wheel odometry~\cite{mandow2007experimental} with online calibration enables a more accurate estimation than visual IMU fusions. 
However, while these approaches were not validated in severely featureless environments, the proposed method was demonstrated to perform highly accurate odometry estimation in such environments by handling degeneracy of the LiDAR point cloud.
Furthermore, these methods do not explicitly estimate the uncertainty of the wheel odometry model as 3D motions.

\section{Methodology}

\subsection{System Overview}
\begin{figure*}[tb]
  \centering
  \includegraphics[width=1.0\linewidth]{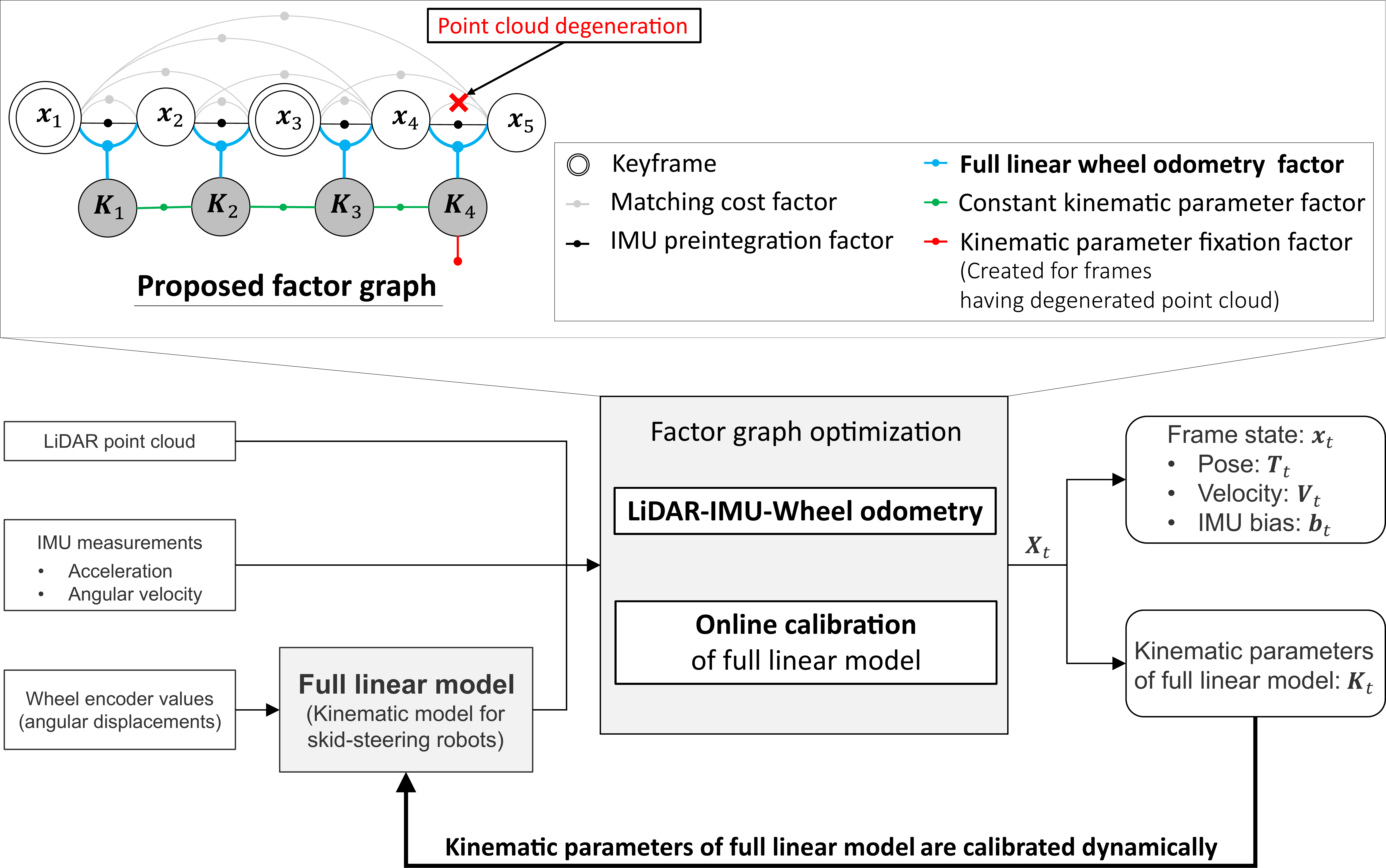}
  \caption{System overview of the proposed method. We jointly solve LiDAR-IMU-wheel odometry and online calibration of full linear model (kinematic model for skid-steering robots) such that all those constraints are consistent. ${\bm x}_t$ is a frame state including IMU pose ${\bm T}_t = [{\bm R}_t | {\bm t}_t] \in SE(3)$, velocity ${\bm v}_t \in \mathbb{R}^3$, and bias ${\bm b}_t = [{\bm b}_t^a, {\bm b}_t^{\omega}] \in \mathbb{R}^6$. ${\bm K}_t \in \mathbb{R}^6$ is the parameters of the full linear model. The proposed method estimates ${\bm x}_t$ and ${\bm K}_t$ in real-time by using measurement sources as in LiDAR point cloud, IMU measurements (acceleration, angular velocity), and wheel encoder values of all wheels.
          }
  \label{fig:factor_graph}
\end{figure*}

To tackle point cloud degeneration and kinematic model errors, we propose a tightly-coupled LiDAR-IMU-wheel odometry incorporating online calibration of a kinematic model for skid-steering robots.
LiDAR-IMU-wheel odometry and online calibration are solved simultaneously in a unified factor graph (FIGURE~\ref{fig:factor_graph}) such that all those constraints are consistent.
The state to be estimated is defined as follows:
\begin{align}
  \label{eq:state_declation}
  {\bm X}_t = [{\bm x}_t, {\bm K}_t], 
\end{align}
\begin{align}
  {\bm x}_t = [{\bm T}_t, {\bm v}_t, {\bm b}_t], 
\end{align}
where ${\bm x}_t$ is a frame state, ${\bm T}_t = [{\bm R}_t | {\bm t}_t] \in SE(3)$ is the IMU pose, ${\bm v}_t \in \mathbb{R}^3$ is the velocity, ${\bm b}_t = [{\bm b}_t^a, {\bm b}_t^{\omega}] \in \mathbb{R}^6$ is the bias of the IMU acceleration and the angular velocity.
${\bm K}_t$ indicates the kinematic parameters of skid-steering robots. 
LiDAR, IMU, and wheel encoders were used as measurement sources to construct the factor graph.
We assume that ${\bm K}_t$ is optimized online along with the sensor states in feature-rich environments before entering environments where LiDAR point clouds can degenerate. 
Once the kinematic model converges to a proper solution, it helps maintain a well-constrained estimation and enables an accurate odometry estimation even in such environments.

Keyframes are a set of frames that have a moderate overlap and help reduce the estimation drift. 
For simplicity, LiDAR point clouds and velocity calculated by the wheel encoders are transformed into the IMU frame as if they are in the same frame. 
Details regarding each factor will be explained in the following subsections.

\subsection{Matching Cost Factor}
The \textit{matching cost factor}~\cite{koide2021globally} is a point cloud-based constraints, which constrains $\bm{T}_i$ and $\bm{T}_j$, such that point cloud $\mathcal{P}_i$ is aligned to the voxelized $\mathcal{P}_j$. 
We use the voxelized GICP (VGICP)-based point cloud registration to constrain many frames in real-time with GPU acceleration~\cite{koide2021voxelized}.
In VGICP, each input point ${\bm p}_k \in \mathcal{P}_i$ is treated as a Gaussian distribution ${\bm p}_k = ({\bm \mu}_k, {\bm C}_k)$, where ${\bm \mu}_k$ and ${\bm C}_k$ are the mean and covariance matrix of the neighbors of ${\bm p}_k$, respectively.
Target points $\mathcal{P}_j$ are discretized by voxelization, and each voxel stores the average of the means and covariance matrices in that voxel. This voxelization enables a quick nearest neighbor search through spatial hashing~\cite{teschner2003optimized}.
In conclusion, the matching cost $e^{\rm {M}}({\bm T}_i, {\bm T}_j)$ is defined as follows:
\begin{align}
  \label{eq:vgicp}
  e^{\rm {M}}({\bm T}_i, {\bm T}_j) &= \sum_{p_k \in \mathcal{P}_i} e^{\text{\rm {D2D}}}({\bm p}_k, {\bm T}_i^{-1} {\bm T}_j), \\
  e^{\text{\rm {D2D}}} ({\bm p}_k, {\bm T}_{ij}) &= {\bm d}_k^\top ({\bm C}'_k + {\bm T}_{ij}{\bm C}_k{\bm T}_{ij}^\top)^{-1} {\bm d}_k,
\end{align}
where $e^{\text{\rm {D2D}}}$ is the distribution-to-distribution distance between an input point ${\bm p}_k$ and the corresponding voxel ${\bm p}_k'~=~({\bm \mu}_k', {\bm C}_k')$, and ${\bm d}_k~=~{\bm \mu}_k' - {\bm T}_{ij} {\bm \mu}_k$ is the residual between ${\bm \mu}_k$ and ${\bm \mu}'_k$. 

We regard the current frame as a new keyframe when an overlap ratio between the point cloud $\mathcal{P}_i$ and keyframes is lower than a threshold~(e.g.,~90\%).
The current frame is constrained by matching cost factors with the last $\it N$ (e.g.,~3) frames and keyframes for reducing the estimation drift, as shown in FIGURE~\ref{fig:factor_graph}.
We carry out the deskewing process for the point cloud $\mathcal{P}_i$ by using IMU measurements before inputting it to the matching cost factor.
Further details can be found in~\cite{koide2022globally}. 

\subsection{IMU Preintegration Factor}
The \textit{IMU preintegration factor} constrains the relative pose and velocity between two consecutive frames (FIGURE~\ref{fig:factor_graph}) by integrating multiple IMU measurements (acceleration ${\bm a}_i$ and angular velocity ${\bm \omega}_i$).
This factor allows for efficient optimization using the preintegration technique eliminating to recompute IMU measurement integration at each optimization iteration.
This factor is vital for efficiently fusing high-frequency IMU measurements and low-frequency sensor measurements (e.g., point clouds). 
The IMU measurements update the sensor state in a time interval $\Delta t$ (i.e., IMU measurements frequency) as follows:
\begin{align}
  \label{eq:imu_evol_R}
  {\bm R}_{t + \Delta t} &= {\bm R}_t \exp \left( \left( {\bm \omega}_t - {\bm b}_t^{\omega} - {\bm \eta}_k^{\omega} \right) \Delta t \right), \\
  \label{eq:imu_evol_v}
  {\bm v}_{t + \Delta t} &= {\bm v}_t + {\bm g} \Delta t + {\bm R}_t \left( {\bm a}_t - {\bm b}_t^a - {\bm \eta}_t^a \right) \Delta t, \\
  \label{eq:imu_evol_p}
  {\bm t}_{t + \Delta t} &= {\bm t}_t + {\bm v}_t \Delta t + \frac{1}{2} {\bm g} \Delta t^2 + \frac{1}{2} {\bm R}_t \left( {\bm a}_t - {\bm b}_t^a - {\bm \eta}_t^a \right) \Delta t^2,
\end{align}
where $\bm g$ is the gravity vector and $\bm{\eta}_t^a$ and $\bm{\eta}_t^{\omega}$ represent the white noise of the IMU measurement. We can calculate a relative sensor motion $\Delta \bm{R}_{ij}$, $\Delta \bm{v}_{ij}$, and $\Delta \bm{t}_{ij}$ by integrating $\Delta \bm{R}_{t+\Delta t}$, $\Delta \bm{v}_{t+\Delta t}$, and $\Delta \bm{t}_{t+\Delta t}$ between time steps $i$ and $j$, respectively. The error $e^{\rm {IMU}}(\bm{x}_i, \bm{x}_j)$ is finally defined as follows:

\begin{align*} 
  e^{\rm {IMU}}\left(\boldsymbol{x}_i, \boldsymbol{x}_j\right) & =\left\|\log \left(\Delta \boldsymbol{R}_{i j}^T \boldsymbol{R}_i^T \boldsymbol{R}_j\right)\right\|^2 \\ & +\left\|\Delta \boldsymbol{t}_{i j}-\boldsymbol{R}_i^T\left(\boldsymbol{t}_j-\boldsymbol{t}_i-\boldsymbol{v} \Delta t_{i j}-\frac{1}{2} \boldsymbol{g} \Delta t_{i j}^2\right)\right\|^2 \\ & +\left\|\Delta \boldsymbol{v}_{i j}-\boldsymbol{R}_i^T\left(\boldsymbol{v}_j-\boldsymbol{v}_i-\boldsymbol{g} \Delta t_{i j}\right)\right\|^2 .
  \stepcounter{equation}\tag{\theequation} 
\end{align*}

Further details can be found in~\cite{forster2016manifold}. 
The IMU preintegration factor makes an estimation robust to rapid motions and degeneracy of point clouds in a short time. 
In addition, this factor reduces the estimation drift in four DoF~\cite{qin2018vins}.

\subsection{Full Linear Wheel Odometry Factor}
\begin{figure}[tb]
  \centering
  \includegraphics[width=0.55\linewidth]{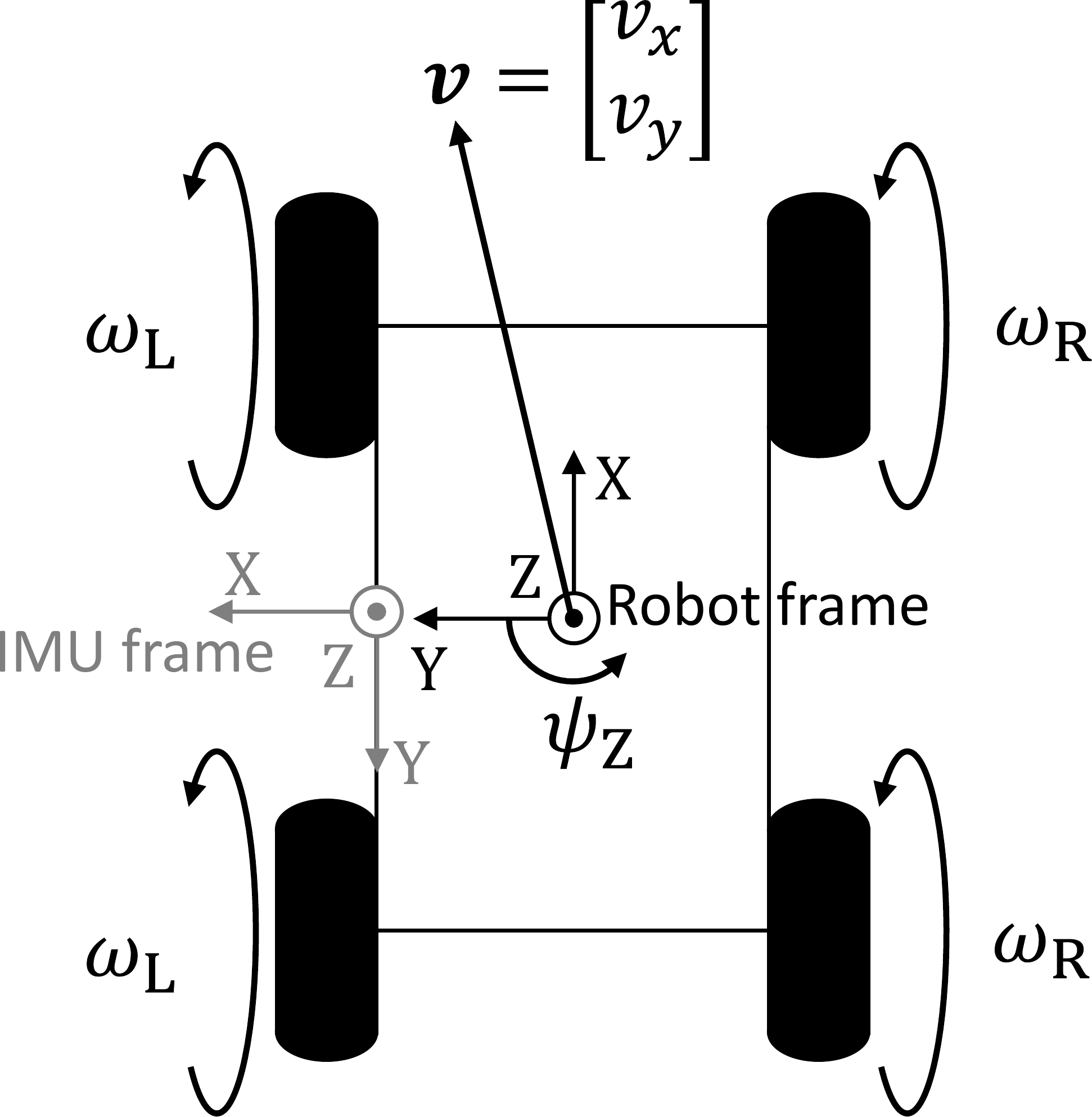}
  \caption{Diagram for kinematics of skid-steering robots. Note that the result of wheel odometry described in the robot frame is finally transformed into the IMU frame.
           The robot velocity $[v_{x}~v_{y}~{\psi_{z}}]^\top$ is calculated by wheel angular velocities \mbox{${\bm \omega} = [\omega_{\rm L} ~\omega_{\rm R}]^\top$} and the $J$ matrix.}
  \label{fig:wheel_robot_description}
\end{figure}
The \textit{full linear wheel odometry factor} constrains consecutive poses and the kinematic parameters; thus this factor performs for motion constrains and online calibration.
To describe a robot motion using wheel encoder values (i.e., wheel odometry estimation), we employed the full linear model~\cite{anousaki2004dead}. 
Compared to other parametric models (e.g.,~\cite{mandow2007experimental, wang2015analysis}), the full linear model exhibits effective characteristics implicitly accounting for the asymmetry of the structure of the robot, mechanical modeling errors, and ground surface changes that are difficult to explicitly model.
Furthermore, Baril~et.al. evaluated these kinematic models~\cite{anousaki2004dead, mandow2007experimental, wang2015analysis} in different environments as in a flat plane and uneven terrain, and the full linear model was the most accurate of all those methods thanks to its flexible expressions~\cite{baril2020evaluation}.

Wheel odometry provides relative transformation and velocity based on kinematic and terrain-dependent parameters, and measurements of wheel encoders. 
For skid-steering robots, the robot velocity $[v_{x}~v_{y}~{\psi_{z}}]^\top$ is calculated as follows:
\begin{align} 
  \begin{bmatrix}
    v_{x} \\
    v_{y} \\
    \psi_{z} \\
  \end{bmatrix}
  = {\bm J}
  \begin{bmatrix}
    \omega_{\rm L} \\
    \omega_{\rm R} \\
  \end{bmatrix},
  \label{eq:wo_for_ssmr}
\end{align}
where $v_{x}$ and $v_{y}$ are the 2D translational velocity in the robot frame, ${\psi_{z}}$ is the angular velocity around the z-axis of the robot frame, and \mbox{${\bm \omega} = [\omega_{\rm L} ~\omega_{\rm R}]^\top$} is the angular velocity of the left and right wheels, as shown in FIGURE~\ref{fig:wheel_robot_description}. ${\bm J}$ is a \(3 \times 2\) matrix that describes the relationship between $[v_{x}~v_{y}~{\psi_{z}}]^\top$ and $\bm \omega$, depends on the kinematic and terrain parameters as follows:
\begin{figure}[tb]
  \centering
  \includegraphics[width=0.8\linewidth]{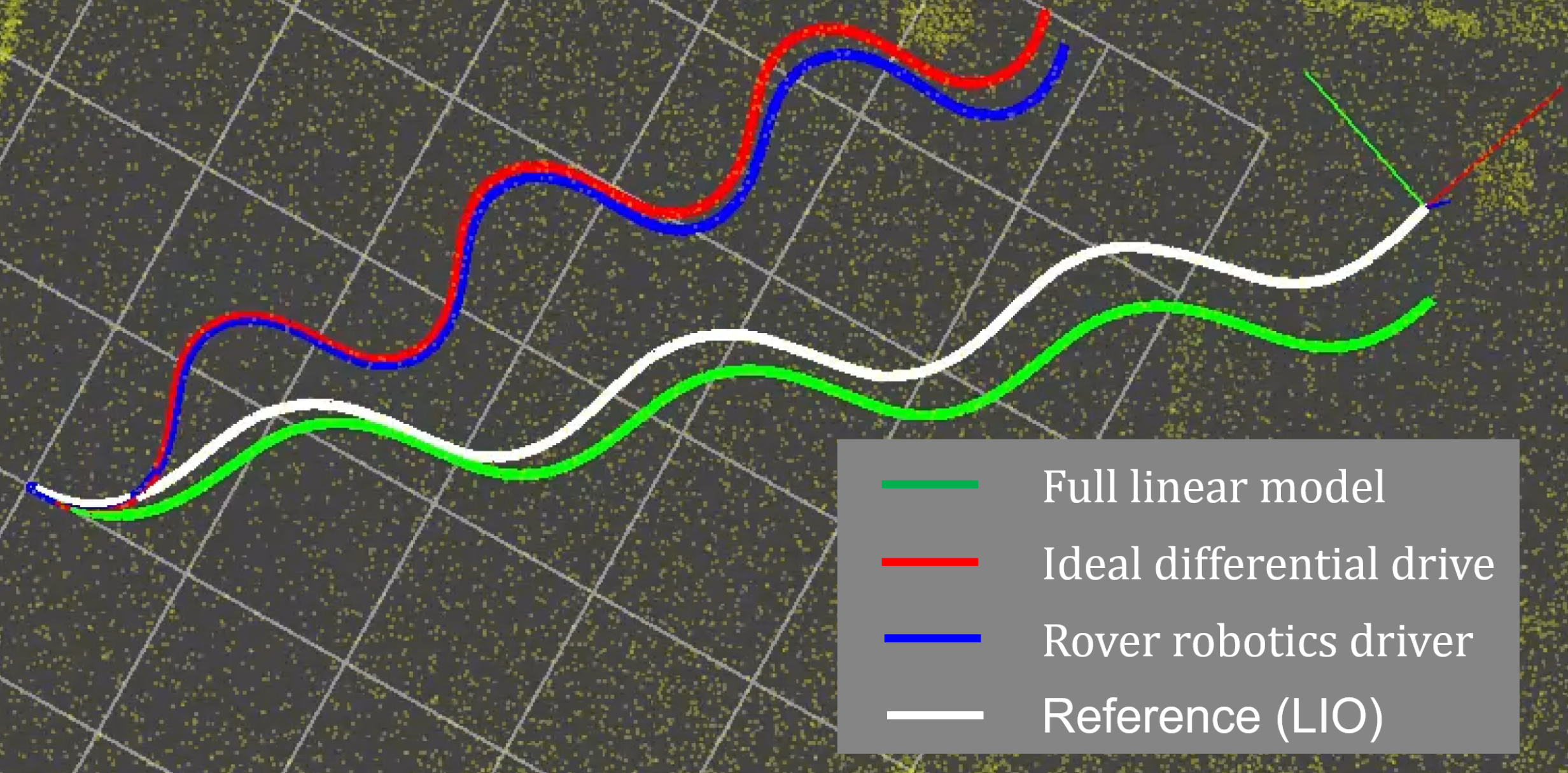}
  \caption{Accuracy validation of the full linear model. We can infer that the official Rover robotics driver uses the ideal differential drive model for expressing the wheel odometry of the skid-steering robot.}
  \label{fig:full_linear_model_validation}
\end{figure}

\begin{align}
  \label{eq:J}
  {\bm J} = \begin{bmatrix}
    J_{11} & J_{12} \\
    J_{21} & J_{22} \\
    J_{31} & J_{32}
  \end{bmatrix}.
\end{align}
The full linear model directly regards these six parameters as state variables; thus ${\bm J}$ is expressed with the parameter vector ${\bm K}$ as follows:
\begin{align}
  \label{eq:JtoK}
  {\bm K} = \begin{bmatrix}
    J_{11} & J_{12} & J_{21} & J_{22} & J_{31} & J_{32}
  \end{bmatrix}^{\top}.
\end{align}

We regard ${\bm K}$ as a time variant value ${\bm K_i}$ as stated in Eq.~\ref{eq:state_declation}.
The full linear wheel odometry factor is a motion constraint depending not only on the robot poses ${\bm T}_i, {\bm T}_j$ but also on ${\bm K_i}$. 
Therefore, ${\bm K_i}$ is calibrated such that all constraints on the factor graph are consistent.
The full linear wheel odometry factor is created between consecutive frame states and the corresponding kinematic parameter, as shown in FIGURE~\ref{fig:factor_graph}.
We then describe an initial value ${\bm K}_0$ for factor graph optimization. 
We use the kinematic parameters of an ideal differential drive model, which does not consider wheel slippage and lateral motion. 
Hence, ${\bm K}_0$ is set as follows:
\begin{align}
  {\bm K}_0 = \begin{bmatrix}
    R/2 & R/2 & 0 & 0 & -R/B & R/B
  \end{bmatrix}^{\top},
  \label{eq:initial_condition_for_calib}
\end{align}
where $R$ and $B$ are nominal values indicating the wheel radius and wheelbase of the robot, respectively.

The displacement $\Delta \bm {o}_{ij} \in se(2)$ is calculated by integrating $[\omega_{\rm L} ~\omega_{\rm R}]^\top$ between the time steps $i$ and $j$ as follows:
\begin{align} 
  \Delta \bm {o}_{ij}
  = {\bm J_i}
  \begin{bmatrix}
    \Delta \theta_{\rm {L},\it{ij}} \\
    \Delta \theta_{\rm {R},\it{ij}} \\
  \end{bmatrix},
\end{align}
where $\Delta \theta_{\rm {L},\it{ij}} = \int_{i}^{j}  \omega_{\rm L}dt$, $\Delta \theta_{{\rm R},\it{ij}} = \int_{i}^{j} \omega_{\rm R}dt$ are the angular displacements of the left and right wheels, respectively. $\bm J_i$ is defined by $\bm K_i$ based on Eq.~\ref{eq:J} and~\ref{eq:JtoK}. We then extend $\Delta \bm {o}_{ij} \in se(2)$ into $\Delta \bm {O}_{ij} \in se(3)$ by setting the translational $z$ and rotational $x$, $y$ elements to zero. Here, $\Delta \bm {O}_{ij}$ is the robot motion described in the robot frame; this motion is treated as the IMU motion $\Delta \bm {\mathcal{O}}_{ij}$ in our implementation. Therefore, $\Delta \bm {O}_{ij}$ is transformed into $\Delta \bm {\mathcal{O}}_{ij}$.
The wheel odometry error ${\bm r}_{ij}^{\rm W}$ is defined as follows:
\begin{align}
  {\bm r}_{ij}^{\rm W} = \log ( {{\bm T}_i^{-1} {\bm T}_j \exp (\Delta \bm {\mathcal{O}}_{ij}) ^{-1}} ).
\end{align}
Consequently, the full linear model-based wheel odometry cost ${\bm e({\bm T}_i, {\bm T}_j, {\bm K}_i)}^{\rm W}$ is defined as follows:
\begin{align}
  {\bm e({\bm T}_i, {\bm T}_j, {\bm K}_i)}^{\rm W} = {{\bm r}_{ij}^{\rm W}}^\top {{\bm C}_{ij}^{\rm W}}^{-1} {\bm r}_{ij}^{\rm W},
\end{align}
where ${{\bm C}_{ij}^{\rm W}}^{-1}$ is the covariance matrix of wheel odometry that is derived as indicated in Section~\ref{subsec:cov_estiamtion_method}.

FIGURE~\ref{fig:full_linear_model_validation} demonstrates an example of the wheel odometry estimation results using the online-calibrated full linear model in comparison to using the conventional ideal differential drive model. 
For reference, the odometry output of the \href{https://github.com/RoverRobotics/roverrobotics_ros2}{rover robotics official driver} and LiDAR-IMU odometry are also shown.
The ideal differential model uses the values of ${\bm K}_0$.
The calibrated full linear model demonstrates a trajectory similar to that of the LiDAR-IMU odometry, and this result shows that this model is the most accurate among these wheel odometry models. 
In addition, we can infer that the official driver is likely computing the wheel odometry similar to that of the ideal differential model. 
\subsection{Constant Kinematic Parameter Factor}
\label{subsec:constant_factor}
A \textit{constant kinematic parameter factor} constraints the consecutive kinematic parameters ${\bm K_i}$ based on a zero-mean Gaussian noise as shown in FIGURE~\ref{fig:factor_graph}.
In the case of approaches related to the IMU preintegration factor, this technique is widely used for modeling random walk bias evolution.
We use this factor for adapting the kinematic parameters to a change in terrain conditions (e.g., a transition from an outdoor to an indoor environment).
In our implementation, we set all diagonal elements of a covariance matrix of this factor to $10^{-10}$.

\subsection{Kinematic Parameter Fixation Factor}
The full linear wheel odometry factor also performs the online calibration for ${\bm K}_i$ in addition to serving as a motion constraint.
Therefore, the online calibration can be unstable if point clouds degenerate.
To avoid this situation, we create a \textit{kinematic parameter fixation factor} for ${\bm K}_i$ as shown in FIGURE~\ref{fig:factor_graph} when the point cloud degeneration is detected on the corresponding frames.
We detect point cloud degeneration by considering the linearized system after the factor graph optimization.
This process consists of the following two steps: 1) the Hessian matrix of a matching cost factor between the latest and last frames is calculated, and 2) if the minimum eigenvalue of the Hessian matrix is smaller than a threshold (i.e., the Hessian matrix is not positive definite), the latest frame is detected as degeneracy. 
If the current point cloud is determined as degeneracy, the kinematic parameter fixation factor for ${\bm K}_i$ is added such that ${\bm K}_i$ becomes those retained just before degeneration occurs to avoid corrupting the optimization.
We set all diagonal elements of a covariance matrix of the kinematic parameter fixation factor to $10^{-10}$ in our implementation.
We also create similar factors for the IMU bias to make optimization stable.

\subsection{Uncertainty Estimation for Wheel Odometry}
\label{subsec:cov_estiamtion_method}
We explicitly estimate the uncertainty (i.e., covariance matrix ${{\bm C}_{ij}^{\rm W}}^{-1}$) for the full linear wheel odometry factor as a 3D motion, not 2D motion.
The uncertainty estimation can make reasonable constraints, particularly for translational z, roll, and pitch motions, which are unobservable for wheel odometry.
Each component of the wheel odometry residual $({r}_{ij, x}^{\rm W}, {r}_{ij, y}^{\rm W}, {r}_{ij, z}^{\rm W}, {r}_{ij, \rm {roll}}^{\rm W}, {r}_{ij, \rm {pitch}}^{\rm W}, {r}_{ij, \rm {yaw}}^{\rm W})$ is assumed to be proportional to the wheel angular displacement $\varDelta \theta_{ij}$:
\begin{align}
  {r}_{ij, x}^{\rm W} &= \varDelta \theta_{ij} a_{j,x} ,\\
  \varDelta \theta_{ij} &= |\varDelta \theta_{\rm {L},\it{ij}}|+|\varDelta \theta_{\rm {R},\it{ij}}|,
\end{align}
where $a_{j,x}$ is a positive proportional coefficient for the translational $x$. We estimate $a_{j,x}$ based on a one-dimensional Kalman filter online. A constant mean model is also assumed for the state transition of the Kalman filter. Therefore, the state and measurement equations are defined as follows:
\begin{align}
  a_{j,x} &= a_{i,x} + q_i,\\
  {r}_{ij, x}^{\rm W} &= \varDelta \theta_{ij} a_{j,x} + s_i,
\end{align}
where $q_i$ and $s_i$ indicate the white noise following the constant process noise $Q$ and observation noise $S$, respectively. According to these equations, $a_{j,x}$ is iteratively updated by the following steps:
\begin{align}
  k_j &= \frac{ \varDelta \theta_{ij} (P_{i} + Q) }{{\varDelta \theta_{ij}}^2 (P_{i} + Q) + S},\\
  a_{j,x} &= a_{i,x} + k_j ({r}_{ij, x}^W - \varDelta \theta_{ij} a_{i,x}),\\
  P_j &= (1 -  \varDelta \theta_{ij} k_j) (P_{i} + Q),
\end{align}
where $k_j$ is the Kalman gain and $P_j$ is the variance of $a_{j,x}$. In our implementation, the initial variance was $P_0=1000$, the initial state was $a_{0,x}=0$, and $Q=10^{-11}$, $S=10^{-3}$. The variance of wheel odometry ${\sigma_{x,ij}}^2$ is calculated as follows:
\begin{align}
  {\sigma_{x,ij}}^2 = (a_{j,x} \varDelta \theta_{ij})^2.
\end{align}

Other than the translational $x$, the variances are computed similarly. 
We assume that each component of wheel odometry is independent; thus, ${{\bm C}_{ij}^{\rm W}}^{-1}$ contains only diagonal components. 
$({\sigma_{x,ij}}^2, {\sigma_{y,ij}}^2, {\sigma_{z,ij}}^2, {\sigma_{\rm {roll},\it {ij}}}^2, {\sigma_{\rm {pitch}, \it{ij}}}^2, {\sigma_{\rm {yaw}, \it{ij}}}^2)$ are set to the corresponding diagonal component in ${{\bm C}_{ij}^{\rm W}}^{-1}$. 
Hence, our method can consider the uncertainties of wheel odometry in six~DoF, not three~DoF.
A constant covariance matrix (e.g, translational elements, rotational elements are set to $3.6\times 10^{-5}$~m$^2$, $2.3\times 10^{-5}$~{rad}$^2$ respectively) is used in the first few steps where the observation is insufficient. 
Specifically, ${{\bm C}_{ij}^{\rm W}}^{-1}$ is estimated when variances of kinematic parameters ${\bm K}_i$ being results of the optimization converge to a value.

\section{Experimental Results}
\begin{figure}[tb]
  \centering
  \includegraphics[width=1.00\linewidth]{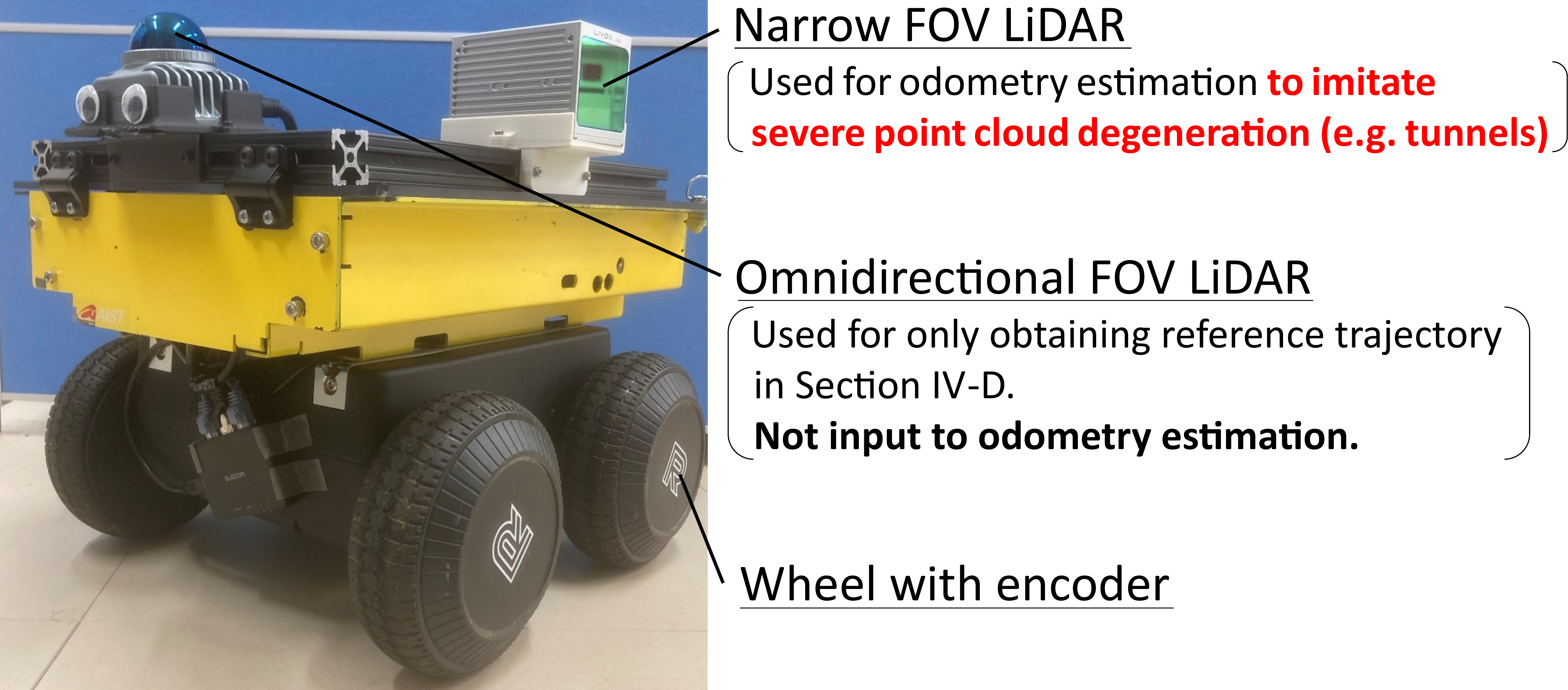}
  \caption{Used testbed for experiments in Section~\ref{subsec:indoor_environment_exp}, \ref{subsec:outdoor_environment_exp}, \ref{subsec:trans_environment_exp}. 
  The narrow FOV LiDAR was placed toward the lateral direction and used for odometry estimation for imitating severe point cloud degeneration (e.g., tunnels).
  Note that the omnidirectional FOV LiDAR was used for only obtaining reference (ground truth of the estimated trajectories) in Section~\ref{subsec:trans_environment_exp}.}
  \label{fig:testbed}
\end{figure}
\subsection{Experimental Setup}
The proposed factor graph was implemented using \href{https://github.com/borglab/gtsam}{GTSAM} and the iSAM2~\cite{kaess2012isam2} optimizer was used to incrementally optimize the factor graph. 
Rover mini (Rover Robotics) was used as a testbed as shown in FIGURE~\ref{fig:testbed}. 
We recorded the point cloud and IMU data using a narrow FOV LiDAR (Livox AVIA) along with the wheel encoder values. 
The narrow FOV LiDAR was used for imitating point cloud degeneration in severe environments, such as tunnels (FIGURE~\ref{fig:exp1_traj_and_experimental_conditions}-(a), (b)).
Point clouds, IMU measurements, and the wheel angular velocities were recorded at 10, 200, and 60 Hz, respectively. 
A velocity command was given to a robot controller for the initial approximately 20 seconds of all the experiments, such that the robot followed a sine curve trajectory for the initial calibration of all elements of the kinematics parameters ${\bm K}_i$.
We assume that ${\bm K}_i$ is calibrated online in feature-rich environments before entering environments where LiDAR point clouds can degenerate, and thus a start location of the robot for all experiments was in a feature-rich environment.

The proposed method was compared with FAST-LIO2~\cite{xu2022fast} and FAST-LIO-Multi-Sensor-Fusion~\cite{zhao2023vehicle} which are state-of-the-art methods.
To conduct ablation studies, the proposed method (Ours) was also compared with two configurations ignoring online calibration and online uncertainty estimation.
We summarize the compared methods as follows.
\begin{itemize}
  \item \textit{Ours w/o online calibration}: Constant kinematics parameters ${\bm K}_0$ derived from the ideal differential drive model (Eq.~\ref{eq:initial_condition_for_calib}) were always set for constructing the wheel odometry-based constraints without online calibration. 
                                              This ideal differential drive model provides nearly the same odometry estimation results as that of the official rover robotics driver (FIGURE~\ref{fig:full_linear_model_validation}); thus, it was considered a reasonable choice. 
  \item \textit{Ours w/o uncertainty estimation}: Constant uncertainties (i.e. covariance matrix) are always set to each full linear wheel odometry factor.
  \item \textit{FAST-LIO2}~\cite{xu2022fast}: FAST-LIO2 is a tightly coupled LiDAR-IMU odometry based on iterated Kalman filter. Although this work can demonstrate accurate odometry estimation in feature-rich environments where point cloud matching works well, this algorithm is not ensured in featureless environments where point clouds degenerate.
  \item \textit{\href{https://github.com/zhh2005757/FAST-LIO-Multi-Sensor-Fusion}{FAST-LIO-Multi-Sensor-Fusion}}~\cite{zhao2023vehicle}: FAST-LIO-Multi-Sensor-Fusion fuses FAST-LIO2~\cite{xu2022fast} and wheel speed-based constraints with online calibration of extrinsic parameters~\cite{zhao2023vehicle}. Details are described in Section~\ref{subsection: senction22}.
\end{itemize}

Refer to the attached video~\footnote{\url{https://youtu.be/Vss86xUhU80}} better to understand the following three experiments of our odometry estimation.

\begin{table}[tb]
  \caption{Ratio of frames encountering degeneracy and absence of the point clouds.}
  \label{tab:points_states}
  \centering
  \scriptsize
  \small
  \scalebox{0.96}{
  \begin{tabular}{ccc}
    \toprule
         &Point cloud degeneracy       & Point cloud absence\\
    \midrule
    Ratio of frames    &  53~\% (1601$/$3029) & 15~\% (447$/$3029)\\
    \bottomrule
  \end{tabular}
  }
\end{table}
\begin{figure}[tb]
  \centering
  \includegraphics[width=0.85\linewidth]{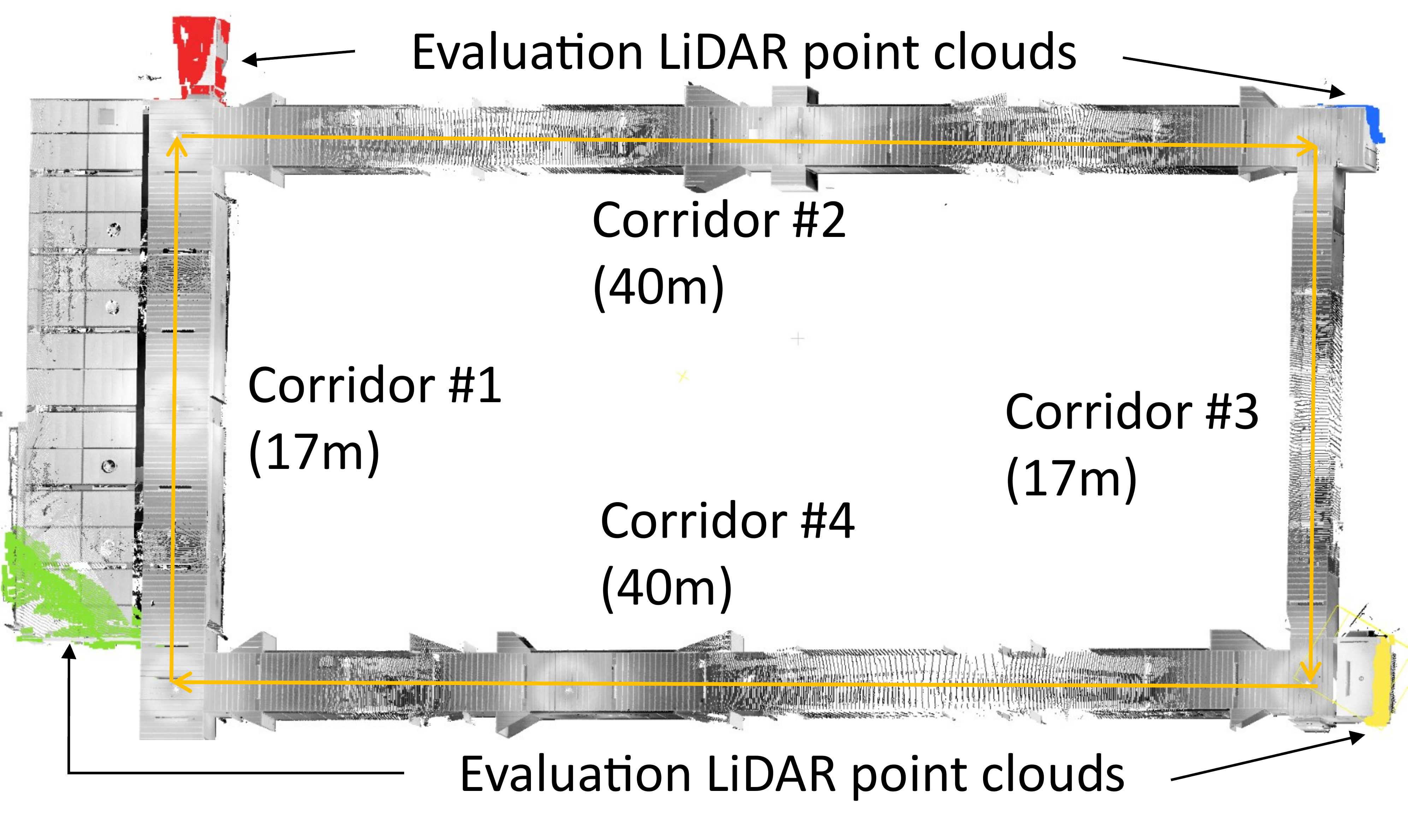}
  \caption{The indoor environment including long corridors, where point cloud degeneration most likely happens (TABLE~\ref{tab:points_states}). 
           The robot traveled about \SI{120}{m}, from the Corridor~\#1 to \#4. 
           \textit{Evaluation LiDAR point clouds} are only used for obtaining the ground truth of the relative sensor position at each corner.}
  \label{fig:prev_map}
\end{figure}
\subsection{Indoor Environment}
\label{subsec:indoor_environment_exp}
The objective of this experiment was to demonstrate that the full linear wheel odometry factor significantly improves the robustness to point cloud degeneracy and kinematic model errors. 
As shown in FIGURE~\ref{fig:prev_map}, the indoor environment consists of long straight corridors, where the almost point cloud degenerates. 
TABLE~\ref{tab:points_states} summarizes the number of frames where the point cloud data was 1)~degenerated, and 2)~unavailable (the number of points was lower than 100). 
Notably, the narrow FOV LiDAR was always pointed at the walls of the corridors in this experiment (FIGURE~\ref{fig:exp1_traj_and_experimental_conditions}~-~(b)); thus, the observed point clouds (FIGURE~\ref{fig:exp1_traj_and_experimental_conditions}~-~(a)) were mostly degenerated (53~\% of all frames). 
Because the LiDAR cannot observe points closer than the minimum observation range (\SI{1}{m}), the point cloud measurements became completely unavailable (15~\% of all frames) when the sensor approached too close to the wall. 

Furthermore, we modified the wheels by attaching styrofoam belts to increase the radius of all the wheels of the robot by 25~\% (FIGURE~\ref{fig:exp1_traj_and_experimental_conditions}~-~(c)). We used ${\bm K}_0$ defined by Eq.~\ref{eq:initial_condition_for_calib} as the initial guess to validate the effectiveness of the online calibration. Ground truth sensor poses were obtained at the corners by manually aligning the \textit{evaluation LiDAR point clouds} into a prior map (FIGURE~\ref{fig:prev_map}). The prior map was obtained by a 3D laser scanner, Focus3D X330 (FARO).

\begin{table*}[tb]
  \caption{Comparison of relative position errors from Corridor~\#1 to \#4 (average and standard deviation of 8 experiments).}
  \label{tab:error_in_indoor_env}
  \centering
  \small
  \begin{tabular}{lcccc}
  \toprule
                         &\textbf{Ours} &Ours w/o online calibration &Ours w/o uncertainty estimation & FAST-LIO2   \\
  \midrule
  \#1 (\SI{17}{m})   & \textbf{0.539 $\pm$ 0.144 m} &  1.176 $\pm$ \SI{0.271}{m} &  0.565 $\pm$ \SI{0.143}{m} & 2.283 $\pm$ \SI{0.558}{m}  \\
  \#2 (\SI{40}{m})   & \textbf{2.188 $\pm$ 0.521 m} &  8.215 $\pm$ \SI{0.600}{m} &  4.900 $\pm$ \SI{0.363}{m} & Error $>$\SI{10}{m}  \\
  \#3 (\SI{40}{m})   & \textbf{0.770 $\pm$ 0.110 m} &  4.907 $\pm$ \SI{0.361}{m} &  1.118 $\pm$ \SI{0.120}{m} & Error $>$\SI{10}{m}  \\
  \#4 (\SI{17}{m})   & \textbf{1.467 $\pm$ 0.331 m} &  9.118 $\pm$ \SI{1.643}{m} &  4.710 $\pm$ \SI{0.454}{m} & Error $>$\SI{10}{m}  \\

  \bottomrule
  \end{tabular}
\end{table*}
\begin{figure}[tb]
  \centering
  \includegraphics[width=0.91\linewidth]{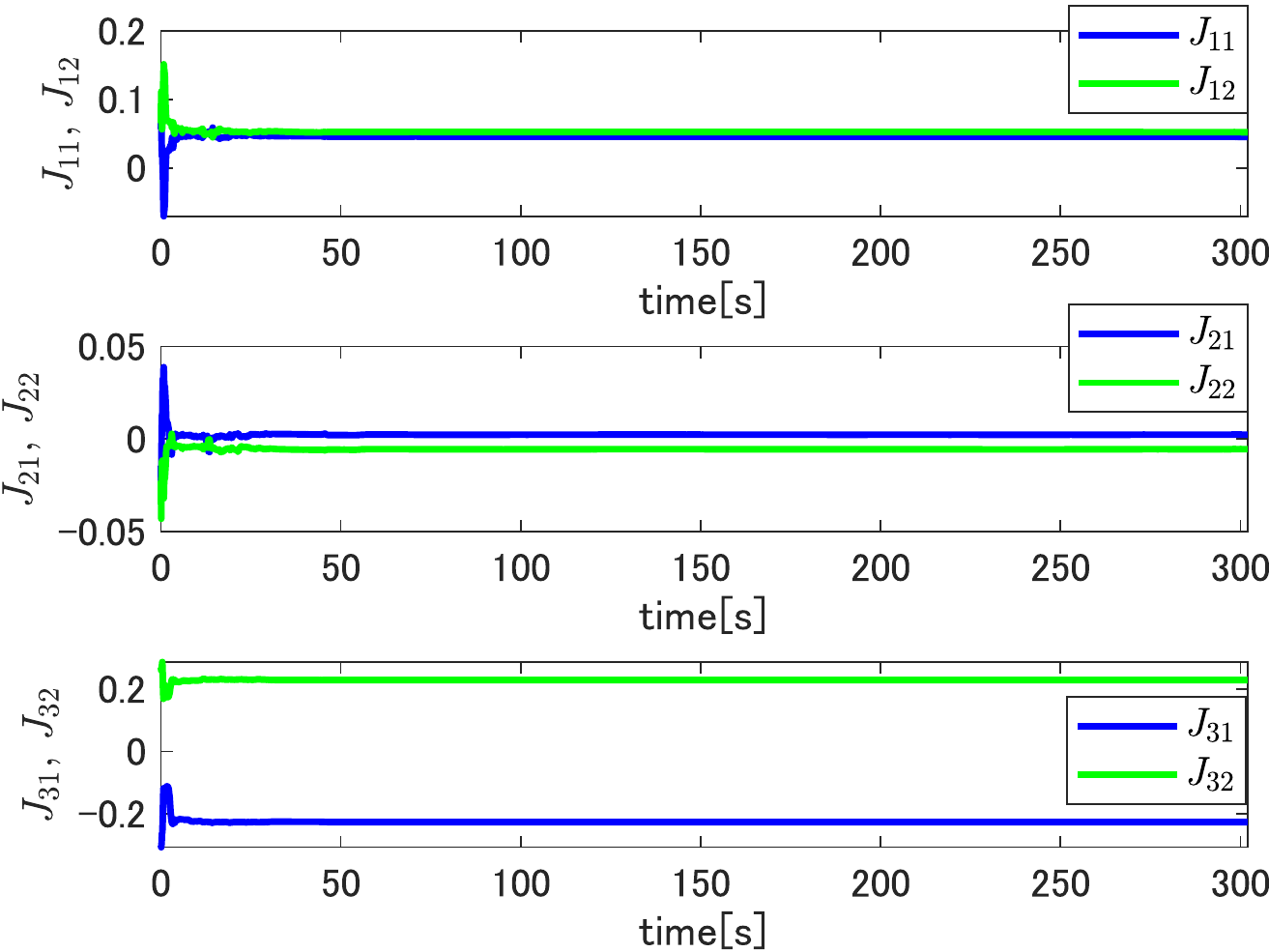}
  \caption{Time histories of the kinematic parameters of the skid-steering robot ${\bm K_i}$.
  ${\bm K_i}$ is few changed after the first 30 seconds because the almost point clouds were degenerated.
  This result indicates that \textit{kinematic parameter fixation factors} worked properly.
  }
  \label{fig:wo_params_result}
\end{figure}
\begin{figure}[tb]
  \centering
  \includegraphics[width=1\linewidth]{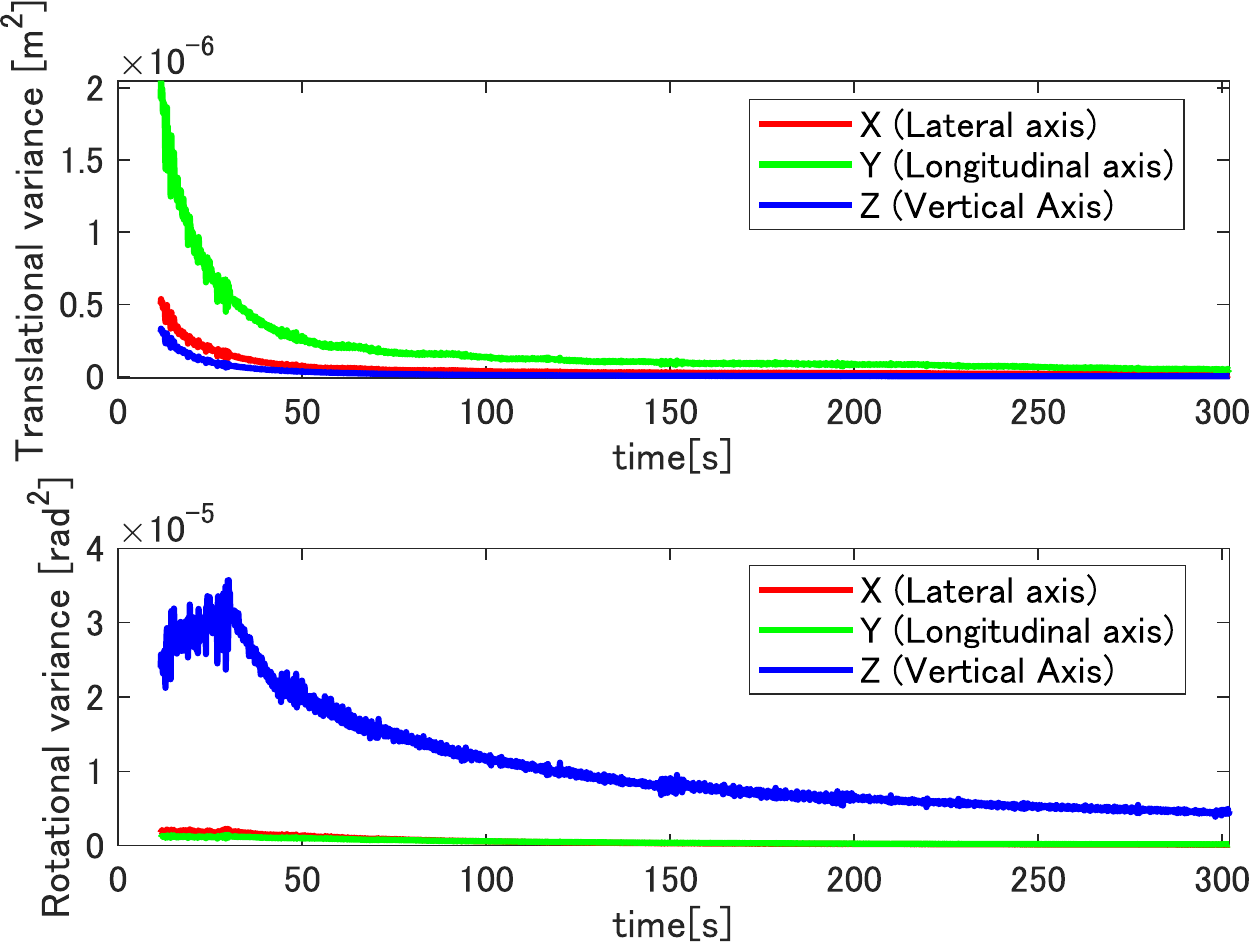}
  \caption{Time histories of the diagonal elements of ${{\bm C}_{ij}^{\rm W}}^{-1}$ in the indoor environment. Note that ${{\bm C}_{ij}^{\rm W}}^{-1}$ is described with respect to the IMU frame as seen in FIGURE~\ref{fig:wheel_robot_description}.}
  \label{fig:wo_cov_result}
\end{figure}

TABLE~\ref{tab:error_in_indoor_env} presents a quantitative analysis of the relative position errors of all methods for Corridor~\#1 to~\#4. 
Accuracy of the proposed method outperforms all other approaches, as shown in FIGURE~\ref{fig:exp1_traj_and_experimental_conditions}~-~(d) and TABLE~\ref{tab:error_in_indoor_env}. 
We can see that both the online kinematics parameter calibration and the uncertainty estimation contributed to the trajectory estimation accuracy.
Although Corridor~\#3 includes intervals where no point cloud data is obtained due to the narrow width of the corridor, odometry estimation remains accurate. 
Therefore, this result validates that the proposed method is robust not only to point cloud degeneracy but also to absence of points.
Also, we can infer that the proposed method has the potential for robustness under other challenging conditions (e.g., rain, snow, and transparent surfaces~\cite{lakmal2020slam}) because the accuracy of the odometry estimation was kept even in the severe scenario where point clouds are completely unavailable.

FIGURE~\ref{fig:wo_params_result} demonstrates that the kinematic parameters ${\bm K_i}$ rapidly converge to reasonable values even though the parameters were initialized with clearly improper values (FIGURE~\ref{fig:exp1_traj_and_experimental_conditions}~-~(c)). 
$J_{11}$ is almost the same as $J_{12}$. 
In addition, $J_{21}$ and $J_{31}$ exhibit approximate symmetry with respect to $J_{22}$ and $J_{32}$, respectively, thus the result of online calibration is reasonable. 
Note that ${\bm K_i}$ is few changed after the first 30 seconds because many \textit{kinematic parameter fixation factors} were created because this indoor environment was the difficult condition almost point clouds degenerate.

FIGURE~\ref{fig:wo_cov_result} is the result of the online estimation of the wheel odometry uncertainties ${{\bm C}_{ij}^{\rm W}}^{-1}$. 
The translational $y$ direction (longitudinal axis) and the yaw direction are larger than the other corresponding elements because the robot was moving on a flat surface.
Whereas, roll and pitch directions of ${{\bm C}_{ij}^{\rm W}}^{-1}$ are quite small consistently; thus this result of uncertainty estimation is reasonable.

\begin{figure}[tb]
  \centering
  \includegraphics[width=1\linewidth]{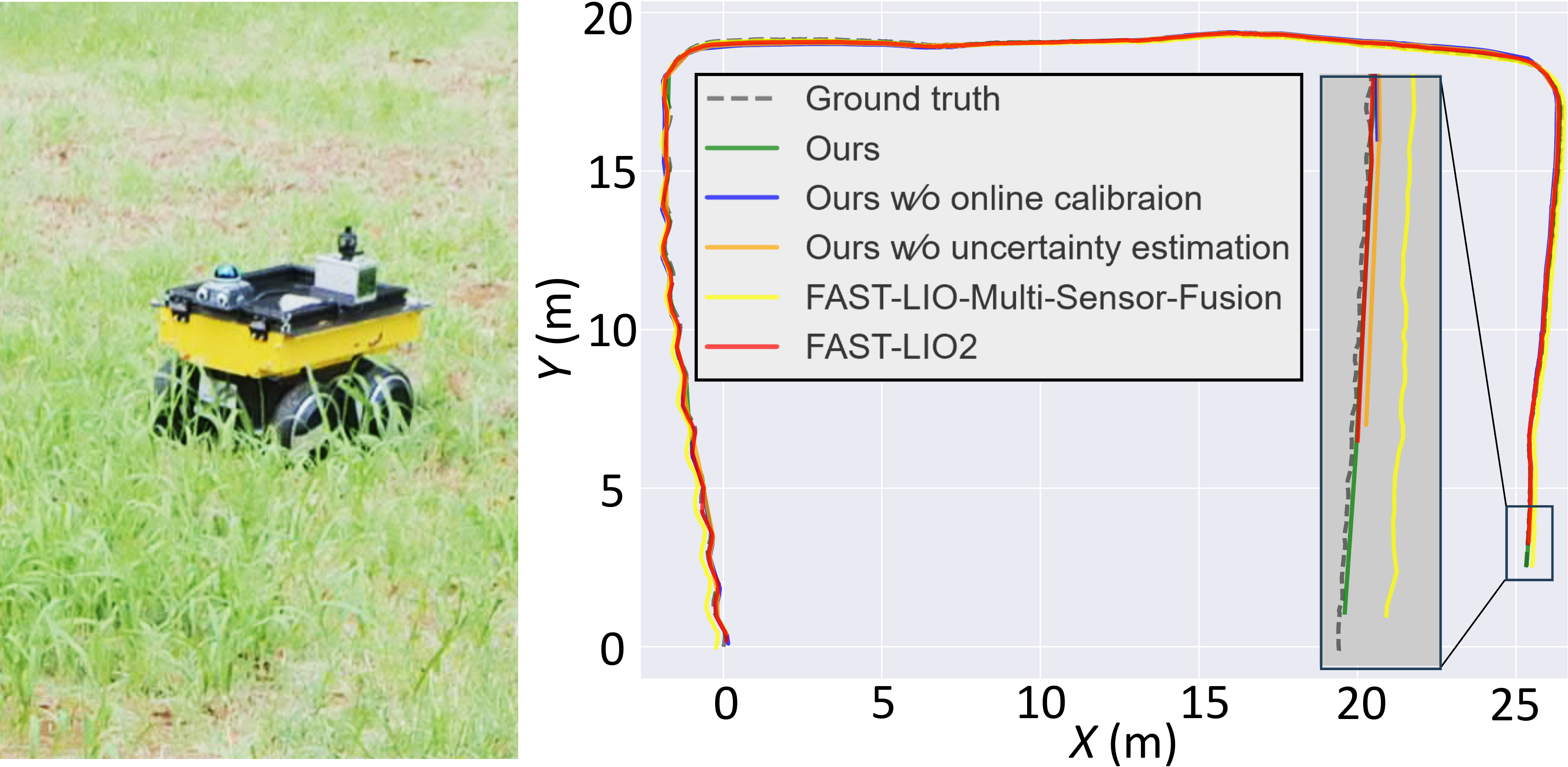}
  \caption{Outdoor environment and trajectory estimation result. This experiment was conducted to verify the online uncertainty estimation in the grass (i.e., rough terrain).}
  \label{fig:exp2_traj}
\end{figure}
\subsection{Outdoor Environment} \label{subsec:outdoor_environment_exp}
The uncertainty estimation of the wheel odometry is important for managing 3D motion fluctuations on uneven terrains. 
To demonstrate the effectiveness and validity of the online uncertainty estimations for such 3D motion fluctuation, the \textit{Ours w/o uncertainty estimation} case is set up with a constant covariance matrix which is obtained by the final values of ${{\bm C}_{ij}^{\rm W}}^{-1}$ estimated in FIGURE~\ref{fig:wo_cov_result} (these values were validated to properly perform in the flat floor by the Section~\ref{subsec:indoor_environment_exp} result).
We obtained the ground truth of the sensor trajectory by using a total station, Trimble S9.
We evaluated the estimation accuracy by using absolute trajectory error (ATE)~\cite{zhang2018tutorial} between estimated trajectories and ground truth.
In TABLE~\ref{tab:exp2_result} and TABLE~\ref{tab:exp3_result}, ATE indicates the global consistency of whole trajectories expressed by the mean and standard deviation of position errors.

\begin{table}[tb]
  \caption{ATE comparison in the grass (rough terrain). The robot traveled about \SI{64}{m}, \SI{186}{s}. ATE shows the global trajectory consistency expressed by the mean and standard deviation of position errors for evaluating estimation results. The proposed method (Ours) is the most accurate of all methods thanks to online calibration and online uncertainty estimation.}
  \label{tab:exp2_result}
  \centering
  \scriptsize
  \small
  \begin{tabular}{lc}
  \toprule
  Methods & ATE [m]\\
  \midrule
  \textbf{Ours} & \bm{$0.049 \pm 0.017$} \\
  Ours w/o online calibration & $0.114 \pm{0.041}$ \\
  Ours w/o uncertainty estimation & $0.092 \pm{0.034}$ \\
  FAST-LIO-Multi-Sensor-Fusion & $0.127 \pm{0.055}$ \\
  FAST-LIO2   & 0.068 $\pm{0.024}$ \\
  \bottomrule
  \end{tabular}
\end{table}
\begin{figure}[tb]
  \centering
  \includegraphics[width=1\linewidth]{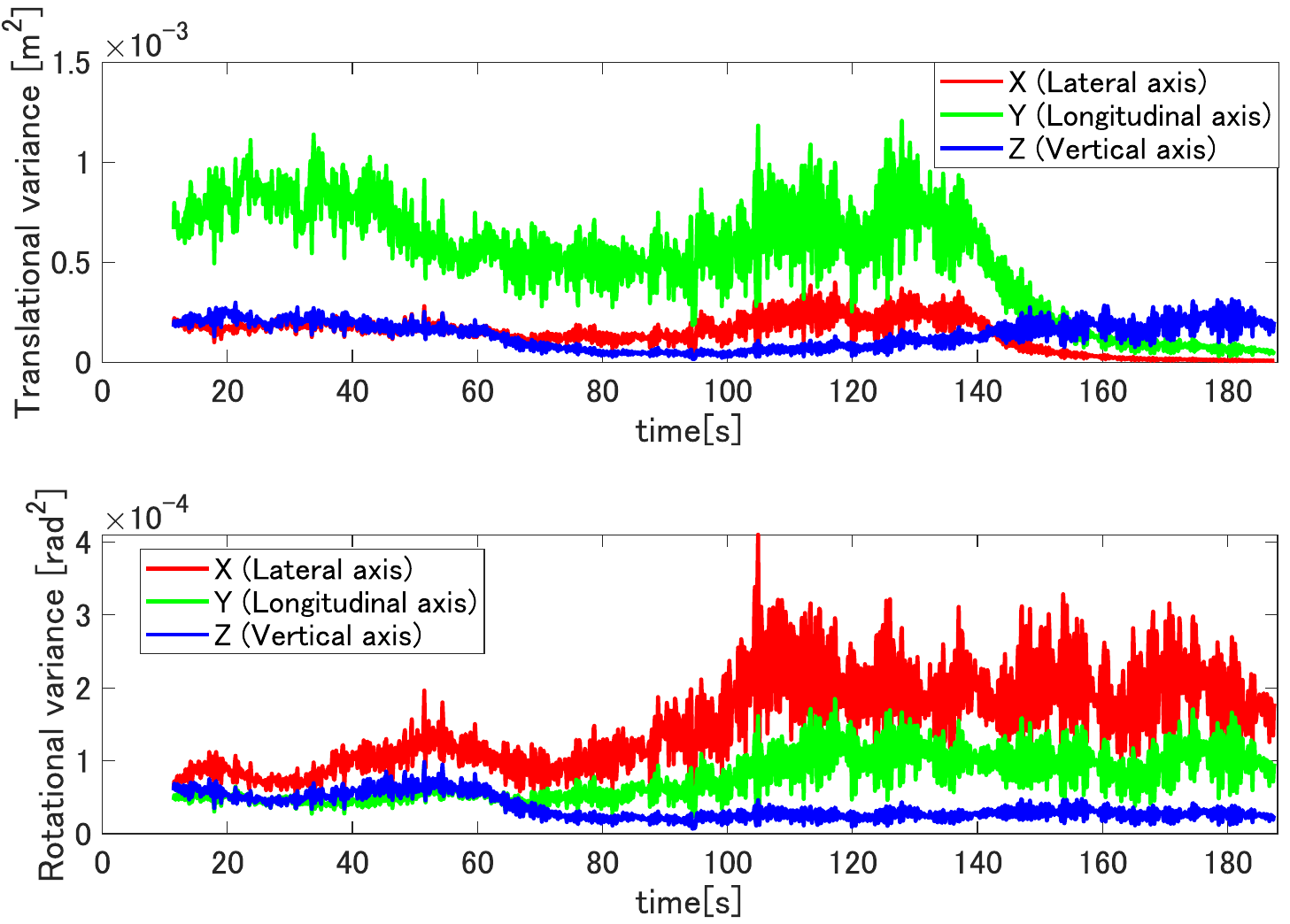}
  \caption{Time histories of the diagonal elements of ${{\bm C}_{ij}^{\rm W}}^{-1}$ in an outdoor environment (grass); the overall components are larger compared to those indoors (FIGURE~\ref{fig:wo_cov_result}).}
  \label{fig:wo_cov_result_for_exp2}
\end{figure}

FIGURE~\ref{fig:exp2_traj} shows the outdoor environment (grass) and the trajectory comparison. 
Owing to the feature-rich environment surrounded by trees and buildings, all trajectories are estimated properly.
TABLE~\ref{tab:exp2_result} indicates absolute trajectory errors (ATEs), and this result shows that the proposed method is the most accurate (ATE: \SI{0.049}{m}) in all approaches thanks to the online calibration and online uncertainty estimation.
FIGURE~\ref{fig:wo_cov_result_for_exp2} demonstrates the results of the wheel odometry uncertainty estimation. 
The overall components are larger compared to the corresponding values for the indoor flat floor (FIGURE~\ref{fig:wo_cov_result}). 
The estimated variances in the roll and pitch directions are particularly large in this environment due to the uneven ground, thus the result of the online uncertainty estimation is reasonable. 
The proposed method (ATE:~\SI{0.049}{m}) is about 2 times as accurate as the \textit{Ours w/o uncertainty estimation} case (ATE:~\SI{0.092}{m}); thus the uncertainty estimations of 3D motion is important for moving on rough terrains.

\begin{figure*}[tb]
  \centering
  \includegraphics[width=1\linewidth]{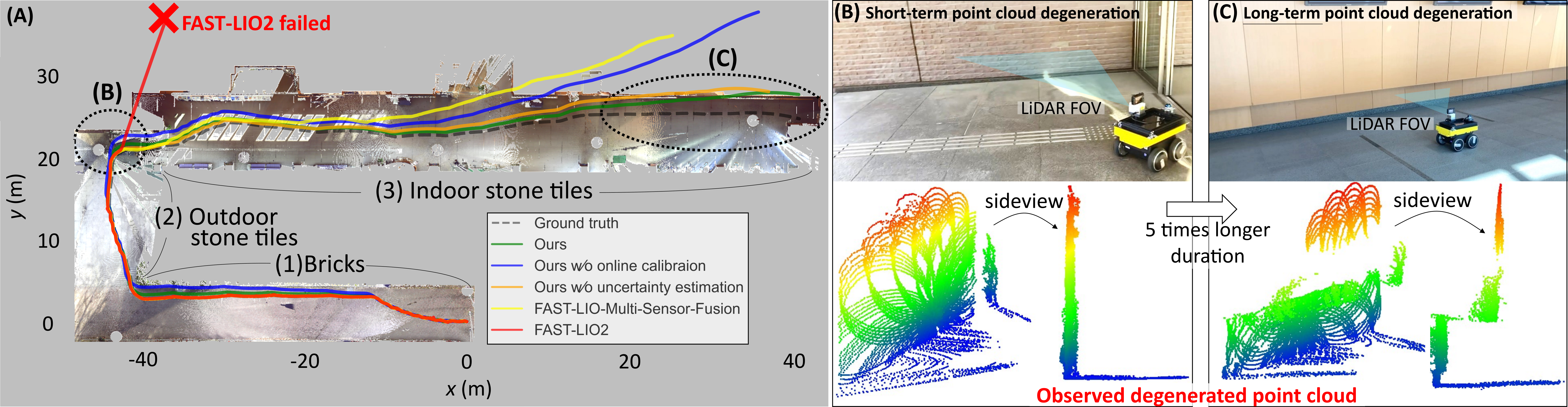}
  \caption{(A)~Odometry estimation results in conditions including point cloud degeneration and terrain condition changes from bricks to outdoor stone tiles, indoor stone tiles.
           (B)~Short-term point cloud degeneration that FAST-LIO2 failed. 
           (C)~Long-term point cloud degeneration that was five times longer duration than the (B) area. 
           As seen in~(B and C), point clouds in these areas are severely degenerated.
           The proposed method (Ours) enabled robust odometry estimation even in long-term point cloud degeneration and terrain condition changes.
           }
  \label{fig:exp3_traj}
\end{figure*}
\begin{figure}[tb]
  \centering
  \includegraphics[width=0.86\linewidth]{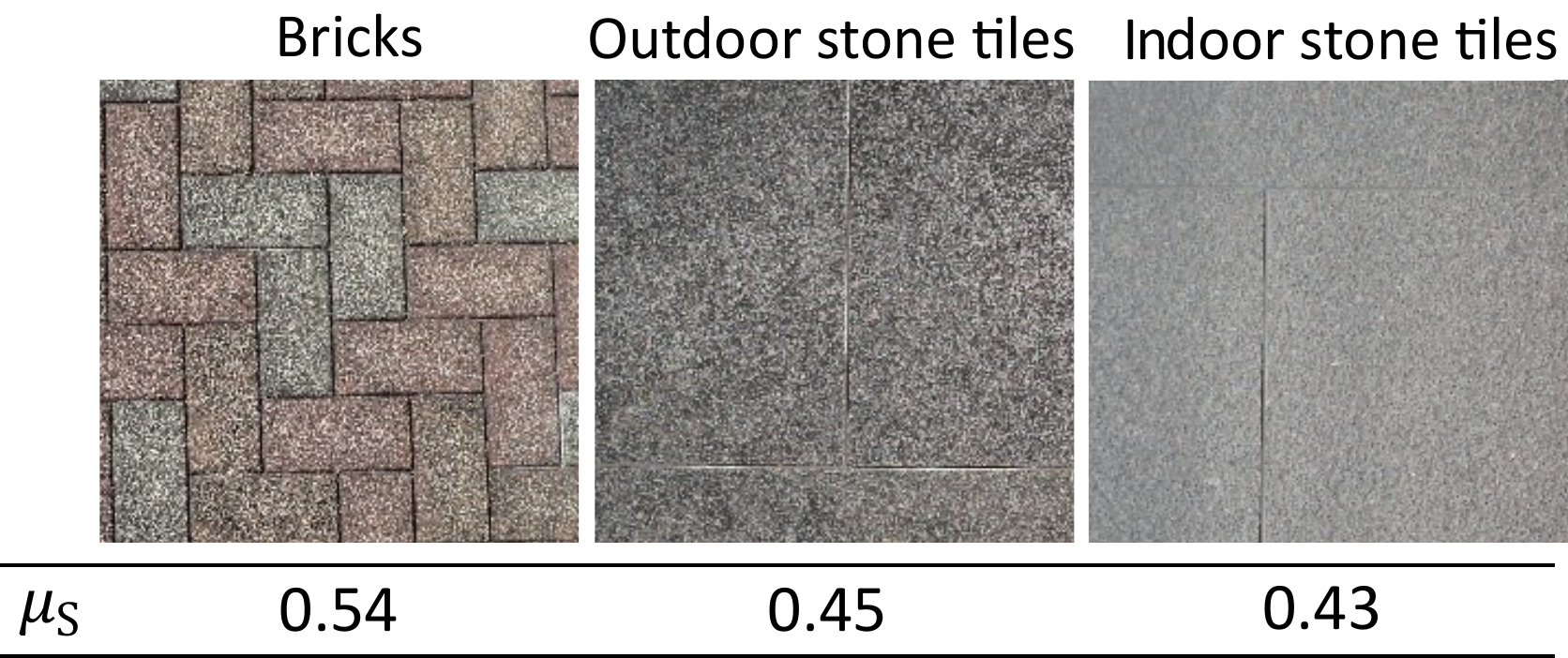}
  \caption{Transition of terrain condition in the third experiment; left: bricks, center: outdoor stone tiles, right: indoor stone tiles. As a reference of the ground characteristic, we measured each condition’s static friction coefficient $\mu _{\rm{s}}$.}
  \label{fig:environments_transition_in_exp3}
\end{figure}

\begin{table}[tb]
  \caption{ATE comparison in the case of transition from bricks to outdoor stone tiles, indoor stone tiles. ATE indicates the global consistency expressed by the mean and standard deviation of position errors for the estimated trajectory. The robot traveled about \SI{144}{m}, \SI{424}{s}.}
  \label{tab:exp3_result}
  \centering
  \scriptsize
  \small
  \begin{tabular}{lc}
  \toprule
  Methods & ATE [m]\\
  \midrule
  \textbf{Ours} & \bm{$0.454 \pm 0.568$} \\
  Ours w/o online calibration & $1.706 \pm{1.025}$ \\
  Ours w/o uncertainty estimation & $0.712 \pm{0.546}$ \\
  FAST-LIO-Multi-Sensor-Fusion~\cite{zhao2023vehicle} & $2.600 \pm{2.106}$ \\
  FAST-LIO2~\cite{xu2022fast}   & Error $>$\SI{10}{m} \\
  \bottomrule
  \end{tabular}
\end{table}
\begin{figure}[tb]
  \centering
  \includegraphics[width=1\linewidth]{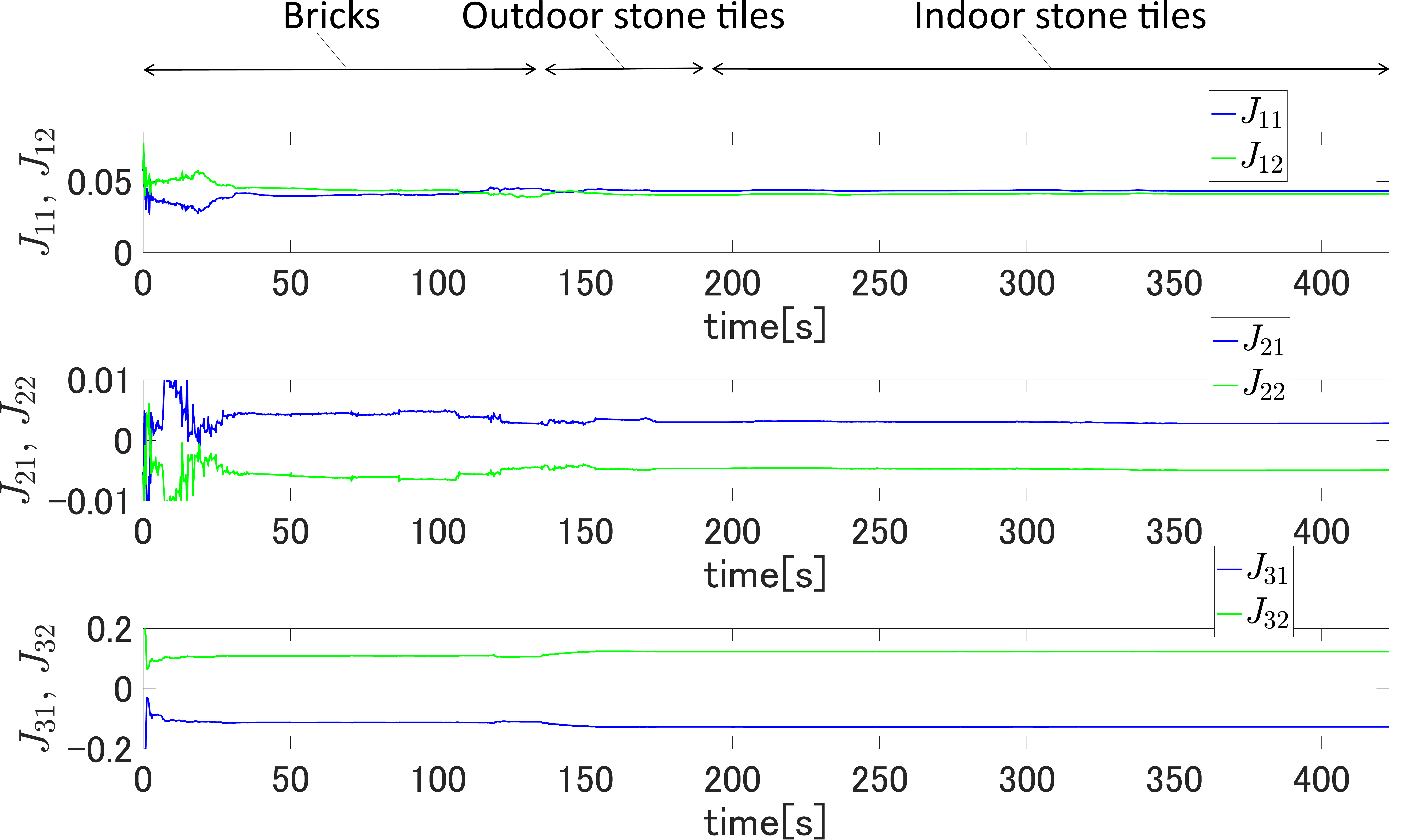}
  \caption{Time histories of ${\bm K_i}$ in the odometry estimation of FIGURE~\ref{fig:exp3_traj}. The process of transition environments is described as follows: 
           bricks (0 to 140 seconds), outdoor stone tiles (140 to 190 seconds), and indoor stone tiles (190 to 420 seconds).
           While ${\bm K_i}$ is adaptively changed when the robot transited from bricks to outdoor door stone tiles, the changes of ${\bm K_i}$ are slight after 190 seconds because the outdoor and indoor stone tiles are almost the same property (FIGURE~\ref{fig:environments_transition_in_exp3}).
           }
  \label{fig:wo_params_result_exp3}
\end{figure}

\subsection{Transition from Outdoor to Indoor Environment} \label{subsec:trans_environment_exp}
The objective of this experiment was to demonstrate that the proposed online calibration enables robust odometry estimation in changing terrain conditions.
In this experiment, the robot transited the following terrain conditions: 1) bricks, 2) outdoor stone tiles, and 3) indoor stone tiles, as shown in FIGURE~\ref{fig:environments_transition_in_exp3}.
The environment included two-point cloud degeneration spots for validating whether kinematic parameter ${\bm K_i}$ is calibrated properly.
In this experiment, point clouds of an omnidirectional FOV LiDAR (Livox MID-360) and its IMU data are recorded in addition to the narrow FOV LiDAR for only estimating a reference trajectory as ground truth (FIGURE~\ref{fig:testbed}).
We manually aligned point clouds of the omnidirectional FOV LiDAR with a prior map (constructed by the same 3D laser scanner as used in Section~\ref{subsec:indoor_environment_exp}). 
The ground truth trajectory was estimated by batch optimization of the scan-to-map registration errors and IMU motion errors.
Note that we input the narrow FOV LiDAR data only into the proposed odometry estimation algorithm.

According to FIGURE~\ref{fig:exp3_traj}-(A) and TABLE~\ref{tab:exp3_result}, the proposed method (ATE:~\SI{0.454}{m}) outperforms all other approaches thanks to online calibration and online uncertainty estimation.
In addition, the mean processing time from making factors to optimizing the factor graph was \SI{0.035}{s} in the third experiment, which has the longest duration in all sequences; thus we validated that the proposed method has real-time capability.
Although the first point cloud degeneration areas in FIGURE~\ref{fig:exp3_traj}-(B) are shorter than the long-term point cloud degeneration area (FIGURE~\ref{fig:exp3_traj}-(C)), FAST-LIO2 failed even in that first area.
While FAST-LIO-Multi-Sensor-Fusion works well in the first short-term point cloud degeneration area thanks to wheel encoder-based constraint, the trajectory of this approach was distorted from the long-term point cloud degeneration areas because only longitudinal velocity was incorporated by the loosely coupling way in contrast to the proposed method.
Even in changing terrain conditions, the proposed method was the most robust to the long-term point cloud degeneration area.
FIGURE~\ref{fig:wo_params_result_exp3} shows a time history of kinematic parameter ${\bm K_i}$ during this experiment.
This result indicates that values of ${\bm K_i}$ are changed after around 140 seconds because the robot transited from bricks to outdoor stone tiles at this time.
In addition, although the robot transited from outdoor stone tiles to indoor stone tiles after around 190 seconds, changes of ${\bm K_i}$ are significantly slight.
As shown in FIGURE~\ref{fig:environments_transition_in_exp3}, the indoor stone's static friction coefficient is the almost same as the outdoor stone's value while these values differ from the brick's value.
Hence, the result of FIGURE~\ref{fig:wo_params_result_exp3} is reasonable.
The reasonable adaptation of ${\bm K_i}$ to the different terrain conditions means the constant kinematic parameter factors worked properly.
Therefore, we validated that the proposed online calibration enables robust odometry estimation even in changing terrain conditions.

\section{Conclusion}
This study presented a tightly-coupled LiDAR-IMU-wheel odometry algorithm with the online calibration of a kinematic model for skid-steering robots. 
In order to perform robust odometry estimation for point cloud degeneration, we proposed the full linear wheel odometry factor and derived its uncertainty. 
The first indoor experiment demonstrated that the proposed method is significantly robust to environments where point clouds degenerate while state-of-the-art LiDAR IMU odometry (FAST-LIO2) fails in these environments. 
In addition, the proposed method estimated the accurate sensor trajectory despite the existence of kinematic model errors (i.e., all-wheel radii are enlarged by 25\%). 
The second outdoor experiment validated that the proposed method can accomplish an accurate odometry estimation in the grass thanks to the uncertainty estimation. 
When the uncertainty estimation was included, the trajectory was twice as accurate compared to not including it. 
The third experiment showed that the kinematic parameters were estimated to adapt to changes in terrain conditions and thus the proposed online calibration performs odometry estimation in robust to terrain conditions changes.
The proposed method outperformed two state-of-the-art approaches (FAST-LIO2, FAST-LIO-Multi-Sensor-Fusion).

In future work, we plan to introduce a nonlinear term for a wheel odometry-based factor to validate a high-speed operation (e.g., large wheel slippage such as drifting) or extremely irregular ground (e.g., highly uneven hill) because these conditions cause a nonlinear motion that the full linear model cannot express accurately.

\bibliographystyle{IEEEtran}
\bibliography{iros2024}

\newpage
\begin{IEEEbiography}[{\includegraphics[width=1in,height=1.25in,clip,keepaspectratio]{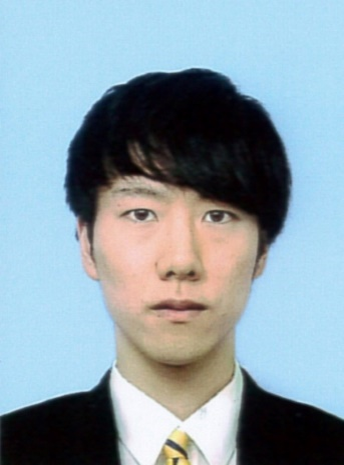}}]{Taku Okawara} received his B.Eng. degree from the Department of Aerospace Engineering from Tokyo Metroporitan University, and M.Eng. degree from the Department of Aerospace Engineering, Graduate School of Engineering, Tohoku University in 2022. He is a Ph.D. candidate in the same department of the master course. He has been also a research assistant at National Institute of Advanced Industrial Science and Technology (AIST) since 2023. His research interests include multi-sensor fusion SLAM for mobile robots, and manipulation for robot arms and climbing robots.
\end{IEEEbiography}

\begin{IEEEbiography}[{\includegraphics[width=1in,height=1.25in,clip,keepaspectratio]{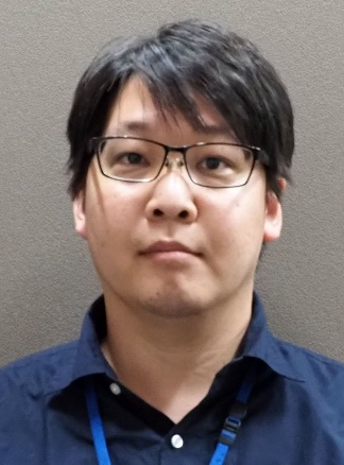}}]{Kenji Koide} received his B.S., M.S., and Ph.D. degree in information engineering from Toyohashi University of Technology in 2013, 2015, and 2019, respectively. In 2018, he joined the Intelligent Autonomous Systems Laboratory, the University of Padova in Italy as a research fellow. Since 2019, he has been working as a research scientist at the National Institute of Advanced Industrial Science and Technology (AIST) in Japan. His research interests include perception and planning for mobile robots.
\end{IEEEbiography}

\begin{IEEEbiography}[{\includegraphics[width=1in,height=1.25in,clip,keepaspectratio]{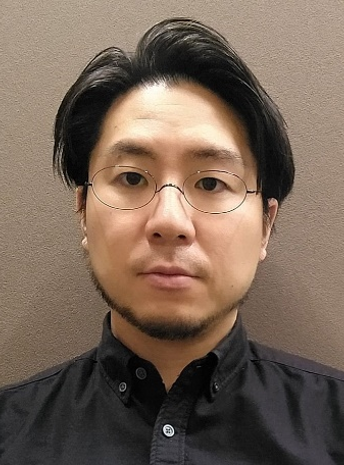}}]{Shuji Oishi} received his M.S. and Ph.D. degree from the Graduate School of Information Science and Electrical Engineering, Kyushu University, in 2012 and 2015, respectively. He joined Department of Computer Science and Engineering in Toyohashi University of Technology as an Assistant professor. Since April 2018, he has been a research scientist at National Institute of Advanced Industrial Science and Technology (AIST). His research interests include 3D modeling/reconstruction with LiDAR(s) / camera(s), visual localization, and autonomous systems based on 3D perception.
\end{IEEEbiography}

\begin{IEEEbiography}[{\includegraphics[width=1in,height=1.25in,clip,keepaspectratio]{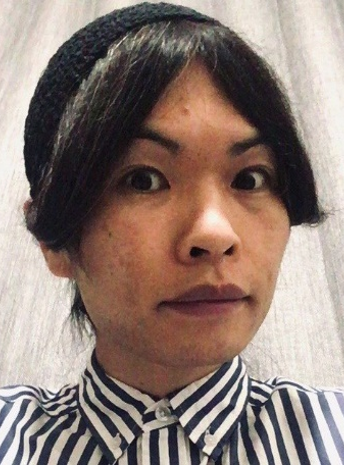}}]{Masashi Yokozuka} received his Ph.D. degree from Utsunomiya University in 2010. Since 2011, he has been a research scientist at the National Institute of Advanced Industrial Science and Technology (AIST) in Japan.
\end{IEEEbiography}

\begin{IEEEbiography}[{\includegraphics[width=1in,height=1.25in,clip,keepaspectratio]{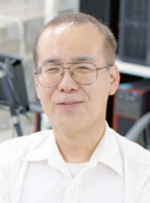}}]{Atsuhiko Banno} received the B.Eng. and M.S. degrees in aeronautics and astronautics from the University of Tokyo in 1994 and 1996, respectively. After leaving the National Research Institute of Police Science, he obtained the Ph.D. in Information Science and Technology from the University of Tokyo in 2006. He had held positions as an assistant professor at the Institute of Industrial Science, the University of Tokyo, and since 2012, he has been working at the National Institute of Advanced Industrial Science and Technology (AIST). His research primarily focuses on 3D reconstruction using cameras and LiDAR. He is a member of the IEEE.
\end{IEEEbiography}
\vspace{-55mm}
\begin{IEEEbiography}[{\includegraphics[width=1in,height=1.25in,clip,keepaspectratio]{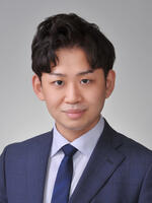}}]{Kentaro Uno}received his B.Eng., M.E., and Ph.D. degrees in Engineering from Tohoku University, Japan. He worked as a software engineering intern at HAKUTO/ispace Inc. from 2018 to 2019, and studied as a research intern at ETH Zurich, Switzerland in 2020. Since 2021, he has been working as an Assistant Professor in the Department of Aerospace Engineering at Tohoku University. His research interests include mobile service robotics in extreme environments such as space and unstructured terrain. 
\end{IEEEbiography}
\vspace{-51mm}
\begin{IEEEbiography}[{\includegraphics[width=1in,height=1.25in,clip,keepaspectratio]{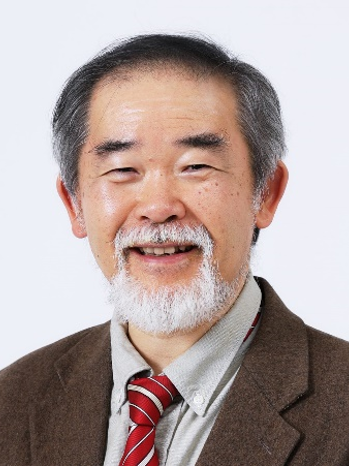}}]{Kazuya Yoshida} received B.E., M.S., and Dr.Eng, degrees in Mechanical Engineering Science from Tokyo Institute of Technology, Japan, in 1984, 1986, and 1990, respectively. Since 2003, he has been Full Professor in Department of Aerospace Engineering at Tohoku University, Japan, leading the Space Robotics Laboratory. He has also taught at International Space University. His research interests include free-flying space robots and lunar/planetary exploration robots. He has been a member of IEEE since 1990.
\end{IEEEbiography}
  
\EOD

\end{document}